\pdfoutput=1

  \RequirePackage{fix-cm}
  \documentclass[twocolumn]{svjour3}          % twocolumn
  \smartqed  % flush right qed marks, e.g. at end of proof
\usepackage{epsfig}
\usepackage{xspace}
\usepackage{color}
\usepackage{multirow}
\usepackage{adjustbox}
\usepackage{subfigure}
\usepackage{amsmath}
\usepackage[noend]{algpseudocode}
\usepackage{algorithmicx,algorithm}
\usepackage{paralist} % compactitem, compactenum
\usepackage{enumitem}
\usepackage{graphicx} %use graph format
\usepackage{epstopdf}% adjust enum margin
\usepackage{booktabs} % Publication quality tables
\usepackage{xspace}
\usepackage{color}
\usepackage{multirow}
\usepackage{graphics}
\usepackage{adjustbox}
\usepackage{subfigure}
\usepackage{amsmath}
\usepackage{paralist} % compactitem, compactenum
\usepackage{enumitem} % adjust enum margin
\usepackage{booktabs} % Publication quality tables
\usepackage{array}

\graphicspath{{figures/}}

\long\def\ignorethis#1{}

\usepackage{tabularx}

\definecolor{gray}{rgb}{0.5,0.5,0.5}
\definecolor{MyBlue}{rgb}{0,0,1.0}
\definecolor{MyYellow}{rgb}{0.9,0.9,0}
\definecolor{MyRed}{rgb}{0.8,0.2,0}
\definecolor{MyGreen}{rgb}{0,0.5,0.0}
\definecolor{MyGray}{rgb}{0.4,0.4,0.4}

\newlength\paramargin
\newlength\figmargin
\newlength\secmargin

\newcolumntype{L}[1]{>{\raggedright\let\newline\\\arraybackslash\hspace{0pt}}m{#1}}
\newcolumntype{C}[1]{>{\centering\let\newline\\\arraybackslash\hspace{0pt}}m{#1}}
\newcolumntype{R}[1]{>{\raggedleft\let\newline\\\arraybackslash\hspace{0pt}}m{#1}}

\def\etal{et~al.\xspace}

\usepackage{url}

  \usepackage{graphicx, ulem, multirow, amssymb, tabularx, paralist, array, tabu, color, subfigure, bbding}
  \usepackage{natbib}

  \makeatletter
  \newcommand{\thickhline}{%
      \noalign {\ifnum 0=`}\fi \hrule height 1pt
      \futurelet \reserved@a \@xhline
  }
  \newcolumntype{"}{@{\hskip\tabcolsep\vrule width 1pt\hskip\tabcolsep}}
  \makeatother
\begin{document}
  
  \title{Learning to Caricature via Semantic Shape Transform
  %MH: no need to follow others in naming. We can come up better terms.
  %\wq{I add the name Semantic-CariGAN}
  }
  
  
  \author{Wenqing Chu$^{1,4}$ \and
  Wei-Chih Hung$^{2}$       \and
          Yi-Hsuan Tsai$^{3}$          \and
          Yu-Ting Chang$^{2}$   \and
          Yijun Li$^{5}$        \and
          Deng Cai$^{1}$        \and
          Ming-Hsuan Yang$^{2}$
  }

  \institute{
            Wenqing Chu \at
              \email{wqchu16@gmail.com}
           \and
           Wei-Chih Hung, Yu-Ting Chang, Ming-Hsuan Yang \at
              \email{\{whung8,ychang39, mhyang\}@ucmerced.edu}
          \and
           Yi-Hsuan Tsai \at
              \email{ytsai@nec-labs.com}
           \and
           Yijun Li \at
              \email{yijli@adobe.com}
           \and
           \Envelope~Deng Cai \at
              \email{dengcai@cad.zju.edu.cn}
           \and
           $^{1}$State Key Lab of CAD\&CG, Zhejiang University, Hangzhou, Zhejiang, China\\
           $^2$Electrical Engineering and Computer Science, University of California, Merced, CA, USA \\
           $^3$NEC Laboratories America, Santa Clara, CA, USA\\ 
           $^4$Tencent Youtu Lab, Shanghai, China\\ 
           $^5$Adobe Research, USA
}
  
  \date{Received: date / Accepted: date}

  \maketitle
  
  \begin{abstract}
 Caricature is an artistic drawing created to abstract or exaggerate facial features of a person.
Rendering visually pleasing caricatures is a difficult task that requires professional skills, and thus it is of great interest to design a method to automatically generate such drawings.
To deal with large shape changes, we propose an algorithm based on a semantic shape transform to produce diverse and plausible shape exaggerations.
Specifically, we predict pixel-wise semantic correspondences and perform image warping on the input photo to achieve dense shape transformation.
We show that the proposed framework is able to render visually pleasing shape exaggerations while maintaining their facial structures.
In addition, our model allows users to manipulate the shape via the semantic map.
We demonstrate the effectiveness of our approach on a large photograph-caricature benchmark dataset with comparisons to the state-of-the-art methods. 
%
%MH: move it to later section
%\red{Code is available at \url{https://github.com/wenqingchu/Semantic-CariGANs}.}

  \end{abstract}

\section{Introduction}
\begin{figure*}[!t]
	\begin{center}
		\centering
			\includegraphics[width= 0.16\textwidth]{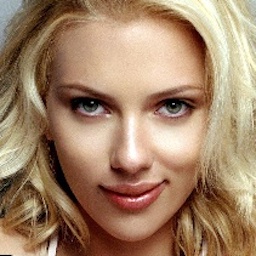}
			\includegraphics[width= 0.16\textwidth]{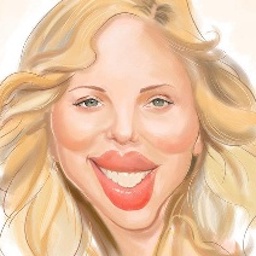} 
			\includegraphics[width= 0.16\textwidth]{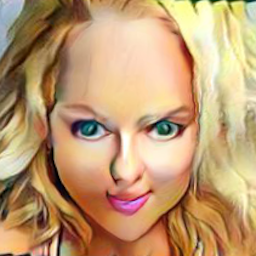} 
			\includegraphics[width= 0.16\textwidth]{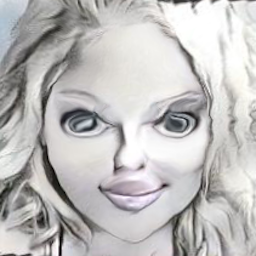} 
			\includegraphics[width= 0.16\textwidth]{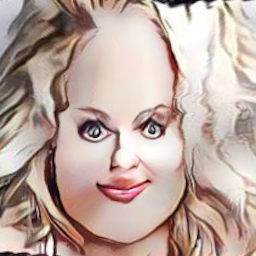} 
			\includegraphics[width= 0.16\textwidth]{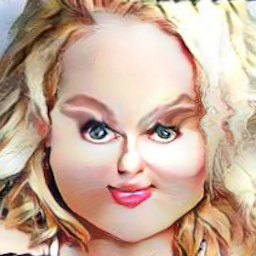} 
			\includegraphics[width= 0.16\textwidth]{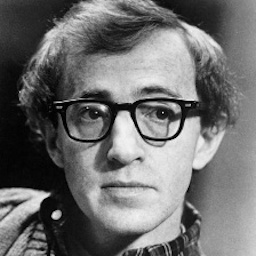} 
			\includegraphics[width= 0.16\textwidth]{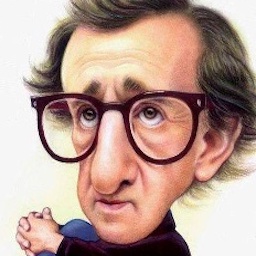} 
			\includegraphics[width= 0.16\textwidth]{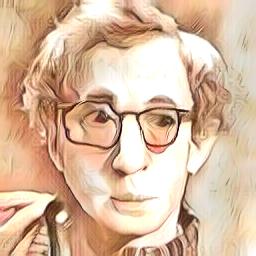} 
			\includegraphics[width= 0.16\textwidth]{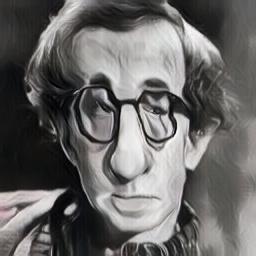} 
			\includegraphics[width= 0.16\textwidth]{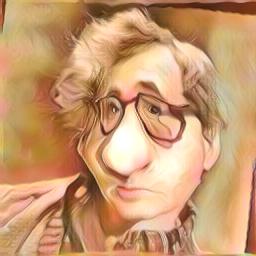} 
			\includegraphics[width= 0.16\textwidth]{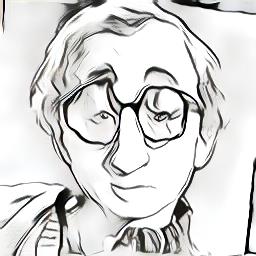} 
		\end{center}
		\vspace{-2mm}
		    \begin{tabular}
		    	{ @{\hspace{10mm}}c@{\hspace{0.65in}} @{\hspace{0mm}}c@{\hspace{1.3in}}
		    		@{\hspace{0mm}}c@{\hspace{0.8in}}
		    	}
		    	Photo & Hand-drawn & Our Results with Diverse Shapes and Styles \\
		    \end{tabular}
		\caption{{\bf Examples of normal photos, hand-drawn caricatures, and a set of caricature outputs generated by the proposed method.} Our approach is able to render a diverse set of visually pleasing caricatures.}
		\label{fig:teaser}
\end{figure*}

Caricature is a rendered image by abstracting or exaggerating certain facial features (e.g., contour, eyes, ear, eyebrows, mouth, and nose) of a person to achieve humorous or sarcastic effects (Fig.~\ref{fig:teaser}).
Caricatures are widely used to depict celebrities or politicians for certain purposes in all kinds of media.
However, generating visually pleasing caricatures with proper shape distortions usually requires professional artistic skills with creative imaginations, which is a challenging task for common users.
Therefore, it is of great interest if caricatures can be generated from normal photos effectively or in a way that users are allowed to flexibly manipulate the output with user controls.

One crucial factor to generate a desirable caricature is to distort facial components properly, i.e., to render personal traits with certain exaggerations. 
Numerous efforts have been made to perform shape exaggeration by computing warping parameters between photos and caricatures from user-defined shapes~\citep{akleman2000making} or hand-crafted rules~\citep{brennan2007caricature,liao2004automatic}.
However, such methods may have limitations on generating diverse and visually pleasing results due to inaccurate shape transformations. 
Recently, image-to-image translation~\citep{isola2017image,zhu2017unpaired,DRIT} and neural style transfer~\citep{gatys2016image,johnson2016perceptual,li2017diversified} algorithms have been developed,
but most techniques can be applied to two domains with local texture variations, not for scenarios where large shape discrepancy exists.

In this work, we aim to create shape exaggerations on standard photos with shape transformations similar to those drawn by artists.
Meanwhile, a rendered caricature should still maintain the facial structure and personal traits.
Different from existing methods that only consider facial landmarks or sparse points \citep{cao2018carigans,shi2019warpgan}, we use a semantic face parsing map, i.e., a dense pixel-wise parsing map, to guide the shape transformation process.
As such, this can provide more accurate mapping for facial details, e.g., shapes of eyebrows, noses, and face contours, to name a few.
Specifically, given an unpaired caricature with a normal photo, we leverage the cycle consistency strategy and an encoder-decoder architecture to model the shape transformation.
Nevertheless, operating this learning process in the image domain may involve noise from unnecessary information in pixels.
Instead, we learn the model directly on the face parsing map, which is the shape transformation of interest.
To learn effective shape transformation, we design a spatial transformer network (STN) to allow larger and flexible shape changes, while a few loss functions are introduced to better maintain facial structures.

To evaluate the proposed framework, we conduct experiments on the photo-caricature benchmark dataset \citep{HuoBMVC2018WebCaricature}.
We perform extensive ablation studies to validate each component of the proposed shape transformation algorithm.
We conduct qualitative and quantitative experiments with user studies to demonstrate that the proposed approach performs favorably against existing image-to-image translation and caricature generation methods.
Furthermore, our model allows users to select the desired caricature semantic shape, with the flexibility to manipulate the parsing map to generate preferred shapes and diverse caricatures.
The main contributions of the paper are as follows:
% itemize
\begin{compactitem}

\item We design a shape transformation model to facilitate the photo-to-caricature generation with visually pleasing shape exaggerations that approach the quality of hand-drawn caricatures.
% %
\item We introduce the face parsing map as the guidance for shape transformation and learn a feature embedding space for face parsing, which allows users to explicitly manipulate the degree of shape changes in the rendered caricature.
% %
\item We evaluate the proposed algorithm on a large caricature benchmark dataset and demonstrate favorable results against existing methods.
\end{compactitem}

\begin{table*}[!t]
	\caption{{\bf Comparisons of caricature generation methods.} 
	Existing approaches mainly leverage the sparse control points for controlling shape exaggerations.
	In contrast, we propose a semantic shape transform method to render visually pleasing and plausible caricatures via using the dense face parsing map. In addition, this allows us to control the shape changes directly on the paring map, while other methods have no control or can only adjust the shape via the one-dimensional warping coefficient.}
	%\vspace{-3mm}
	\label{table:comparison_caricature}
	\scriptsize
    \resizebox{\textwidth}{!}{
	\newcommand{\tabincell}[2]{\begin{tabular}{@{}#1@{}}#2\end{tabular}}
	\newcolumntype{P}[1]{>{\centering\arraybackslash}p{#1}}
	
	\centering
	%\begin{tabular}{m{2.25cm}  P{1.8cm} P{1.8cm} P{1.8cm} P{2.8cm} P{2.8cm}  P{2.1cm}}
	\begin{tabular}{cccccc}
		\toprule
		Method & Akleman \etal~\cite{akleman2000making} & Brennan \etal~\cite{brennan2007caricature} & WarpGAN~\cite{shi2019warpgan} &  CariGANs~\cite{cao2018carigans} & Ours \\
		\midrule
		\scriptsize{Input Requirement} & \scriptsize{Photo}  & \scriptsize{Line Drawing} & \scriptsize{Photo} &  \scriptsize{Photo} & \scriptsize{Photo} \\
		\scriptsize{Transformation Space} & \scriptsize{Landmark Points} & \scriptsize{None} & \scriptsize{Control Points} & \scriptsize{Landmark Points} & \scriptsize{Face Parsing Map} \\
		\scriptsize{Output Shape Control} & \scriptsize{User-defined}  & \scriptsize{Rules} & \scriptsize{Warping Coefficient}  & \scriptsize{Warping Coefficient} & \scriptsize{Face Parsing Map} \\
		%\midrule
		\bottomrule
	\end{tabular}
	}
	\vspace{-5mm}
\end{table*}

\section{Related Work}
%We discuss the recent work on image translation, caricature generation and spatial transformer networks.

%\vspace{-3mm}
%\paragraph{Image Translation.}
{\flushleft \bf Image Translation.}
Numerous methods based on GANs \citep{goodfellow2014generative} have been recently developed for image translation for paired \citep{isola2017image,chang2018pairedcyclegan}, unpaired \citep{zhu2017unpaired,kim2017learning,liu2017unsupervised}, and multimodal \citep{zhu2017toward,DRIT,huang2018multimodal} settings.
While most GANs based methods require large image sets for training, neural style transfer models \citep{gatys2015texture,gatys2016image} only need a single style image as the reference. 
%
% The neural style transfer algorithm by Gatys \etal~\cite{gatys2015texture,gatys2016image} optimizes a style loss based on the discrepancy of the Gram matrices of deep features extracted from the content and style images. 
%
A number of methods have since been developed~\citep{johnson2016perceptual,li2017diversified,chen2017stylebank,liao2017visual,li2017demystifying} to improve style translation or runtime performance.
In the photo-to-caricature task, however, neither GAN-based approaches or neural style transfer methods take the large shape discrepancy across domains into account. 
Unlike existing image translation methods, our algorithm enables shape exaggerations by utilizing an encoder-decoder architecture with the guidance of a face parsing map to densely model the shape transformation.

\begin{figure*}[!t]
	\begin{center}
	%\fbox{\rule{0pt}{3in} %\rule{0.9\linewidth}{0pt}}
	\includegraphics[scale=0.5]{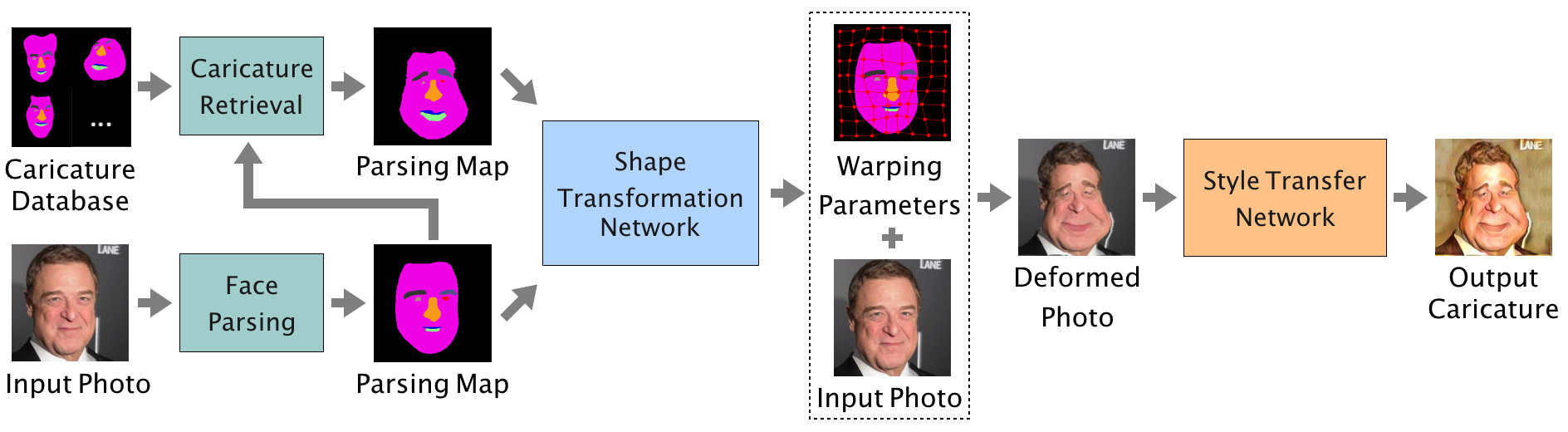}
	\end{center}
	\vspace{-5mm}
	\caption{{\bf Overall framework of the proposed caricature generation method.}
% 	\wq {
	Given an input photo, we first obtain its face parsing map and then retrieve caricature parsing maps from a large-scale database.
% 	} 
	Second, we feed these two maps into the proposed shape transformation network to predict the warping parameters and produce the deformed photo. 
	Third, we utilize a style transfer network to generate the final output with caricatured textures.
% 	Here we omit to present the style reference for simplicity but focus on the shape part.
	}
	\label{fig:framework}
	\vspace{-3mm}
\end{figure*}

%HERE
%\vspace{-3mm}
%\paragraph{Caricature Rendering.}
{\flushleft \bf Caricature Rendering.}
Learning to caricature from photos is mainly concerned with modeling shape transformation.
However, it has not been widely explored due to the large domain gap between photos and caricatures.
Some early methods \citep{akleman1997making,akleman2000making} rely on user-defined source and target shapes to compute the warping parameters, but the rendered images do not exhibit caricature styles. 
Several rule-based methods~\citep{luo2002exaggeration,liao2004automatic,brennan2007caricature} are developed to perform shape exaggeration for each facial component. 

Recently, CycleGAN-based image translation methods~\citep{li2018carigan,zheng2017photo} have been proposed for caricature generation with facial landmarks as conditional constraints.
However, the geometric structures of the images generated with these algorithms are still close to the original photos, with limited exaggerated effects. 
To increase shape exaggeration, \cite{cao2018carigans} use a geometric exaggeration model by leveraging landmark positions to predict key points in the subspace formed by principal components. 
The WarpGAN model~\citep{shi2019warpgan} learns to directly predict a set of control points used to warp the input photo based on unpaired adversarial learning.
Despite showing promising results, their shape exaggerations are still limited due to the use of sparse landmarks or control points.
In contrast, we use a pixel-wise parsing map and model shape transformation, thereby generating plausible exaggerations while retaining the facial traits of the input image.
Furthermore, due to the learned representations in the parsing space, users can explicitly manipulate the facial map to generate preferred shapes. 
In Table~\ref{table:comparison_caricature}, we compare our algorithm with existing methods in terms of shape transformation and requirements.

%\vspace{-3mm}
%\paragraph{Spatial Transformer Network.}
{\flushleft \bf Spatial Transformer Network.}
The spatial transformer networks (STNs)~\citep{jaderberg2015spatial} are developed to improve object recognition performance by reducing input geometric variations. 
Numerous variants have since been developed for a wide range of computer vision applications~\citep{wu2017recursive,zhou2018gridface,Wang_afrcnnCVPR2017,dai2017deformable,ganin2016deepwarp,park2017transformation,shu2018deforming} that require geometric constraints. 
Close to our work is the method proposed by~\cite{lin2018stgan} in which low-dimensional warping parameters are learned to manipulate foreground objects for image composition.  
In our task, we also introduce an STN to predict warping parameters to enable shape exaggeration on normal photos.
In contrast, we need denser and more complex deformations instead of low-dimensional affine transformation~\citep{jaderberg2015spatial} or homography transformation~\citep{lin2018stgan}. 
Furthermore, our shape transformation network leverages the facial parsing maps as an additional input to focus on the semantic facial structure.
 \section{Algorithmic Overview}
\label{sec:overview}
In this section, we introduce the overall framework of the proposed caricature generation method (see Fig.~\ref{fig:framework}).

%\vspace{-3mm}
%\paragraph{Model Inputs.}
{\flushleft \bf Model Inputs.}
Existing methods~\citep{zheng2017photo,li2018carigan,cao2018carigans} use images or sparse landmarks as the inputs to capture facial structure information for shape exaggeration.
However, these approaches are not effective for capturing large shape transformations in caricatures as their appearances of facial components are significantly different from the ones in normal photos.
In this paper, we use the facial parsing map which provides dense semantic information to facilitate computing correspondences between faces of the photo and caricature.
%
%MH: is this the best place to cite this paper?
In practice, we adopt the adapted parsing model in~\citep{chu2019weakly} to account for the domain shift issue. 
More details about the training strategy and network architecture of the parsing model can be found in~\citep{chu2019weakly}. 
%
% \wq{
Note that during training the parsing network, only 17 facial landmarks of caricatures are provided like other caricature generation methods~\citep{cao2018carigans,li2018carigan}.
During the testing stage, we do not need any key-point annotations.
Although the parsing quality could not be always satisfactory (e.g., IoU is 86.5\% on facial skins), we design loss functions (in Section \ref{sec:loss}) to make the deformed shape smooth and plausible.
%

\begin{figure}[!t]
	\footnotesize
	\centering
% 	\renewcommand{\tabcolsep}{1pt} % adjust horizontal space
% 	\renewcommand{\arraystretch}{0.6} % adjust vertical space
% 	\begin{tabular}{cccc}

	\includegraphics[width=0.85\linewidth]{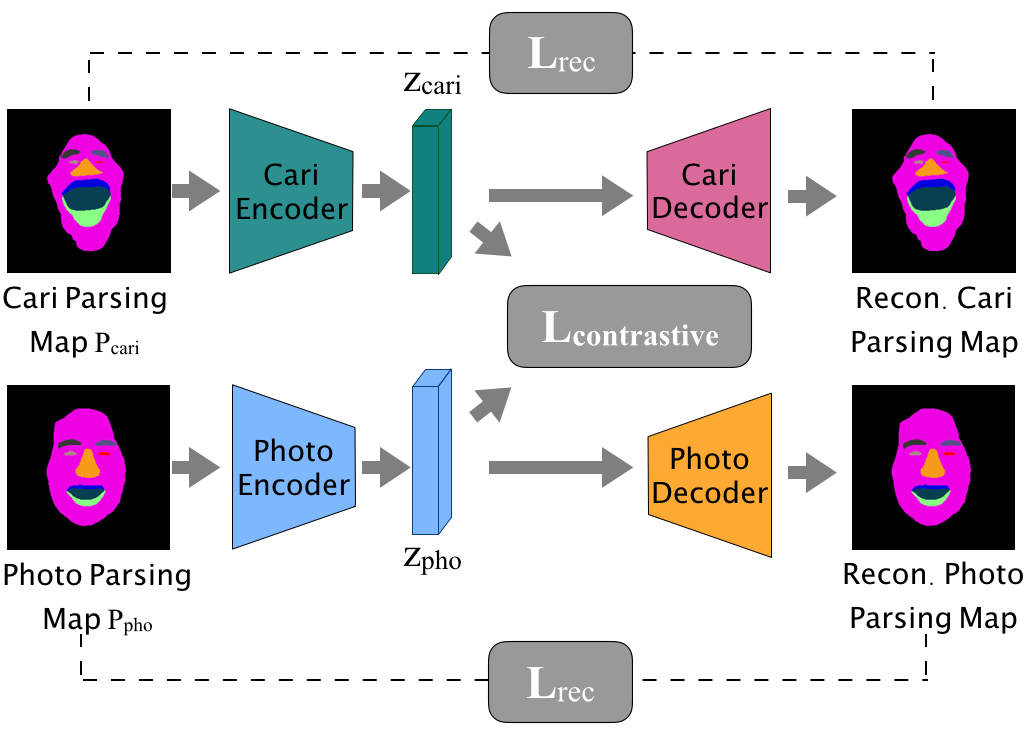} 
		
% 	\end{tabular}
	\vspace{-2mm}
	\caption{{\bf Framework of caricature retrieval}. Our goal of training the retrieval model is to learn the photo and caricature embedding on parsing maps, so that given the photo parsing map, the model is able to retrieve proper caricature maps during testing.}
% 	\yh{move this figure to the later section}
	\label{fig:retrieve} %% label for entire figure
	\vspace{-3mm}
\end{figure}

%\vspace{-3mm}
%\paragraph{Caricature Retrieval.}
%\red{
{\flushleft \bf Caricature Retrieval.}
We note that the selection of target semantic shapes is useful and important for diverse caricature generation.
In this work, we develop a parsing map retrieval model with the large-scale dataset WebCaricautre~\citep{HuoBMVC2018WebCaricature} as the gallery images.
%
%\red{
Given a photo parsing map $\mathbf{P}_{pho}$, we aim to retrieve suitable caricatures $\mathbf{P}_{cari}$ as inputs to calculate the following shape transformation.
The key challenge is how to find the appropriate embedding space to perform retrieval.
Here, we assume that if $\mathbf{P}_{pho}$ and $\mathbf{P}_{cari}$ belong to the same identity, $\mathbf{P}_{cari}$ could be a good reference.
Thus, they should be close to each other in the embedding space.
As shown in Fig.~\ref{fig:retrieve}, we utilize the contrastive loss $\mathcal{L}_{contrastive}$~\citep{chopra2005learning} to enforce the caricature and photo embeddings of the same person being close to each other, while the reconstruction loss $\mathcal{L}_{rec}$ helps preserve the content of the parsing maps through minimizing the Euclidean distance between the input parsing maps and reconstructed ones.
%}
%

%\red{
The encoder consists of four basic dense blocks \citep{huang2017densely} and a flatten layer to obtain a global $128$-dimensional vector $z_{cari}$/$z_{pho}$, while the decoder is composed of four symmetric dense blocks with transposed convolution layers.
The contrastive loss for positive and negative pairs is defined as:
\begin{equation}
\mathcal{L}_{contrastive}=
\begin{cases}
\|z_{cari} - z_{pho}\|_{2},& \mbox{positive,}\\
\max(m-\|z_{cari} - z_{pho}\|_{2},0),& \mbox{negative,}
\end{cases}
\end{equation}
where the hyper-parameter $m$ is the margin set to $2$ in this work.
%}
%\wq{ add more descriptions for the an-notations in Fig.3}
%
During testing, we pre-compute the caricature embedding of our gallery caricatures in the training set, and use the photo encoder to compute the embedding of the testing photo.
To find multiple caricature parsing maps, we first retrieve the top 5 caricature embeddings that are closest to the photo one based on the Euclidean distance, and then use their associated caricature parsing maps as our final outputs.
%
%During training, we utilize the Adam~\citep{kingma2014adam} optimizer and set the batch size as $32$.
%
%We set the initial learning rate as $0.001$ and fix it for the first 100 epochs, and linearly decay the learning rate for another 100 epochs.
%More details are presented in the appendix.

%\vspace{-3mm}
%\paragraph{Overall Pipeline.}
{\flushleft \bf Overall Pipeline.}
Given the input photo, we first use a caricature retrieval model to automatically recommend the proper caricature parsing map, along with the photo parsing map as inputs to the next phase.
Second, to better mimic the process of drawing caricatures, we decompose the caricature generation pipeline into two stages: shape transformation and style transfer. 
In the first stage, we propose a semantic-aware shape transformation network to learn dense warping parameters that enable shape exaggerations for the input photo.
Once obtaining the image with the deformed shapes on facial components, we use a reference based feed-forward style transfer network \citep{huang2017arbitrary} to perform the photo-to-caricature texture translation and obtain the final caricature output.

 \begin{figure*}[!t]
	\begin{center}
	%\fbox{\rule{0pt}{3in} \rule{0.9\linewidth}{0pt}}
	\includegraphics[scale=0.45]{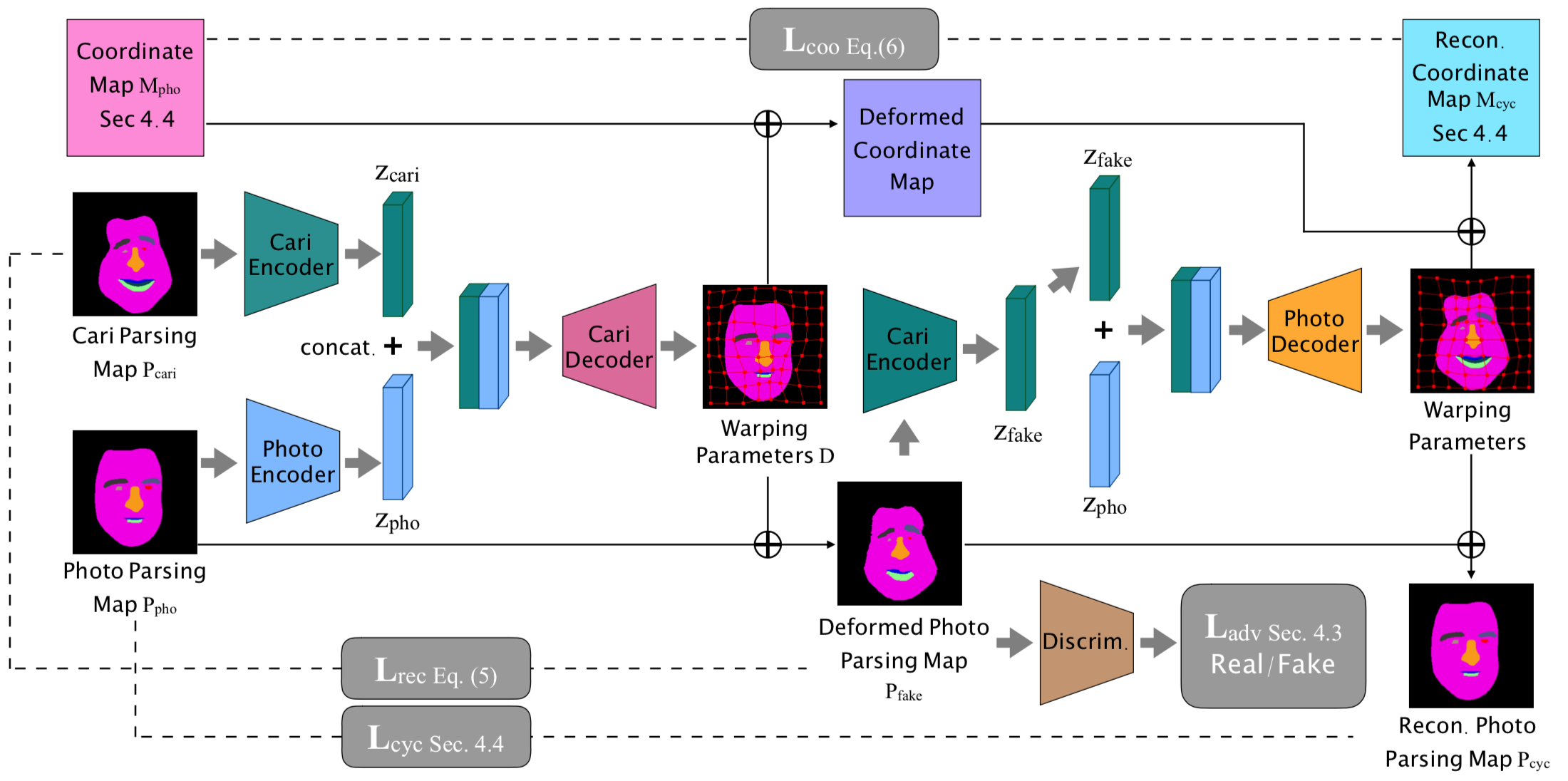}
	\end{center}
	\vspace{-3mm}
	\caption{{\bf Proposed shape transformation network.} We first feed face parsing maps to each encoder to extract latent feature encodings $z$. Second, we concatenate two features as the input of the decoder to predict the warping parameters $D$ and thus produce the parsing map $\mathbf{P}_{fake}$ of the deformed photo. A reconstruction loss $\mathcal{L}_{rec}$ is then computed between $\mathbf{P}_{pho}$ and the parsing map of the caricature $\mathbf{P}_{cari}$.
	To ensure local details and cycle consistency, we further incorporate an adversarial loss $\mathcal{L}_{adv}$ on $\mathbf{P}_{fake}$ and a cycle consistency loss $\mathcal{L}_{cyc}$ between $\mathbf{P}_{pho}$ and the reconstructed parsing map of the photo $\mathbf{P}_{cyc}$.
	In addition, we add a coordinate-based loss $\mathcal{L}_{coo}$ to constrain the alignment based on pixel locations.
	}
	\label{fig:shape_network}
	\vspace{-3mm}
\end{figure*}

\section{Semantic Shape Transformation}
Given a portrait photo and a recommended caricature, our algorithm transforms the portrait photo to have a similar facial structure to the recommended caricature.
In contrast to most image translation methods that learn pixel mapping in the image space, we learn the dense pixel correspondence between inputs and outputs on facial components through the face parsing map.
We use an encoder-decoder architecture, where the encoder extracts feature representations of the parsing map and the decoder composed of an STN module estimates the warping parameters, i.e., the transformation from the parsing map of the photo to the caricature one.
It is worth noting that learning in the semantic space is easier than the original image space due to less appearance discrepancy.
The overall architecture and designed loss functions are presented in Fig.~\ref{fig:shape_network}.

\subsection{Encoder}
Here, we describe how to obtain compact representations for the facial structures of caricatures and photos.
%
% To capture the facial structure information, we utilize the face parsing maps as inputs to avoid the interference from appearances.
We first denote the face parsing maps of the photo and caricature as $\mathbf{P}_{pho}$ and $\mathbf{P}_{cari} \in \mathbb{R}^{C \times H \times W}$, where the image height and weight are denoted as $H$ and $W$, and $C$ is the number of the facial component category. 
As a result, in each channel, there is a binary map to describe each facial component.
Considering the distributions of facial structures between photos and caricatures are quite different, we use two independent encoders in which each network consists of several dense blocks.
%
% The encoder network consists of several dense blocks and 
As such, the encoded feature is a compact 128-dimensional vector. 
In the following, we denote them as $z_{pho}$ and $z_{cari}$. 

%HERE
\subsection{Decoder}

Once obtaining the latent feature $z$ from the encoder to represent the facial structure, the goal of the decoder is to predict the dense correspondence denoted by a tensor $\mathbf{D} \in \mathbb{R}^{2 \times H \times W}$.
Specifically, ($\mathbf{D}_{1,i,j}$, $\mathbf{D}_{2,i,j}$) indicates the corresponding target position when warping each pixel ($i,j$) from the recommended caricature to the input photo.
To perform warping, we first concatenate the latent encoding $z_{pho}$ and $z_{cari}$, and then feed this feature to the decoder.
Here, we introduce a spatial transformer network module to generate the 2D shape transformation parameters for each pixel.
%
% to generate the 2D shape transformation parameters for each pixel.
As a result, we can apply the differentiable bilinear sampling operation to the photo parsing map $\mathbf{P}_{pho}$ and obtain the deformed photo parsing map $\mathbf{P}_{fake}$.
%
% And the deformed photo will be generated in the same way and further processed for producing desired caricature.

The ensuing question is how to enforce the generated $\mathbf{P}_{fake}$ to resemble the parsing map of the real caricature.
To this end, we impose three constraints on $\mathbf{P}_{fake}$: 1) the generated parsing map should be densely reconstructed with respect to the recommended one in the semantic space; 2) a cycle consistency is measured between the generated parsing map and the recommended one; 3) a coordinate-based reconstruction is used to regularize alignment at locations of facial components.
In the following, we introduce the loss functions based on the above-mentioned constraints. 
%

%HERE
\subsection{Reconstruction in the Semantic Space}
\label{sec:loss}
%\paragraph{Reconstruction Loss.}
{\flushleft \bf Reconstruction Loss.}
First, a natural way to enforce similarity is to require $\mathbf{P}_{fake}$ and $\mathbf{P}_{cari}$ identical at each pixel.
Therefore, we minimize the L1 distance between them as below:
\begin{equation}
    \mathcal{L}_{rec}^{p} = \frac{1}{C \times H \times W} \sum_{i=1}^{C} \sum_{j=1}^{H} \sum_{k=1}^{W} \|\mathbf{P}_{cari}^{(i,j,k)} - \mathbf{P}_{fake}^{(i,j,k)}\|_{1}.
\end{equation}
However, we find that this function is not effective to reconstruct every facial component. 
For instance, face skin regions can have a large overlap between $\mathbf{P}_{cari}$ and $\mathbf{P}_{fake}$, while smaller components such as eyes do not usually have spatial overlaps.
To handle this issue, we design a location-aware metric which measures the distance of the center in the same facial component between $\mathbf{P}_{cari}$ and $\mathbf{P}_{fake}$. 
We average the locations of each pixel to obtain a mean location ($x^c, y^c$) for each facial component $c$ and obtain:
\begin{equation}
\left\{
\begin{aligned}
    & \mathbf{x}^{c} = \frac{1}{H \times W} \sum_{j=1}^{H} \sum_{k=1}^{W} \mathbf{P}^{(c,j,k)} \times j. \\
    & \mathbf{y}^{c} = \frac{1}{H \times W} \sum_{j=1}^{H} \sum_{k=1}^{W} \mathbf{P}^{(c,j,k)} \times k. \\
\end{aligned}
\right.
\end{equation}
This location-aware reconstruction loss is defined by:
\begin{equation}
    \mathcal{L}_{rec}^{l} = \frac{1}{C} \sum_{i=1}^C (\|\mathbf{x}_{fake}^{i} - \mathbf{x}_{cari}^{i}\|_{2} + \|\mathbf{y}_{fake}^{i} - \mathbf{y}_{cari}^{i}\|_{2}).
\end{equation}
Furthermore, we define a global alignment loss by matching the number of pixels in each facial component:
\begin{equation}
    \mathcal{L}_{rec}^{n} = \frac{1}{C } \sum_{i=1}^{C} \|(\sum_{j=1}^{H} \sum_{k=1}^{W} \mathbf{P}_{cari}^{(i,j,k)}) - (\sum_{j=1}^{H} \sum_{k=1}^{W} \mathbf{P}_{fake}^{(i,j,k)})\|_{2}.
\end{equation}
The full objective function for reconstruction is:
\begin{equation}
    \mathcal{L}_{rec} = \mathcal{L}_{rec}^{p} + \lambda_{l} \mathcal{L}_{rec}^{l} + \lambda_{n} \mathcal{L}_{rec}^{n},
    \label{eq:rec}
\end{equation}
where the hyper-parameters $\lambda_{l}$ and $\lambda_{n}$ control the importance of each term.
In this work, we use $\lambda_{l} = \lambda_{n} = 2$ in all experiments.
To encourage our loss functions to pay attention to small components, we introduce a weight $\lambda_{comp}^c$ adaptively computed as the reciprocal pixel-ratio in each facial component $c$ of the entire image, i.e., $\mathcal{L}_{rec} = \sum_{c=1}^{C} \lambda_{comp}^c \mathcal{L}_{rec}^c$, where $\mathcal{L}_{rec}^c$ indicates the loss $\mathcal{L}_{rec}$ for each semantic category $c$.
%
%HERE
%\vspace{-3mm}
%\paragraph{Adversarial Loss.}
{\flushleft \bf Adversarial Loss.}
The reconstruction-based loss can be used to recover the global structure.
On the other hand, the GAN-based adversarial loss~\citep{goodfellow2014generative} has been shown to be effective for preserving local details.
In this work, we adopt a similar approach based on the GAN loss but in the semantic parsing space.
To ensure the generated $\mathbf{P}_{fake}$ to look like a realistic caricature, we employ adversarial learning to match the distribution of $\mathbf{P}_{fake}$ to the one of caricatures, i.e., $\mathbf{P}_{cari}$.
We adopt the same training scheme and loss function as the Wasserstein GAN model~\citep{arjovsky2017wasserstein}.
We denote this adversarial loss for the generator as $\mathcal{L}_{adv}$.
%

%HERE
\subsection{Cycle Consistency}
Similar to the CycleGAN model~\citep{zhu2017unpaired}, we add the cycle consistency to make the transformation more stable.
Specifically, we first input $\mathbf{P}_{fake}$ to the same caricature encoder and extract the feature $z_{fake}$.
We then concatenate $z_{fake}$ and $z_{pho}$, feed them into a decoder and recover the original face parsing map of the photo, denoted as $\mathbf{P}_{cyc}$.
Here, we utilize the same reconstruction-based loss as in \eqref{eq:rec}, defined as $\mathcal{L}_{cyc}$, in which the only difference is that we compute the loss between $\mathbf{P}_{cyc}$ and $\mathbf{P}_{pho}$.

\begin{figure*}[!t]
	
	\footnotesize
	%	\tiny
	\centering
	\renewcommand{\tabcolsep}{1pt} % adjust horizontal space
	\renewcommand{\arraystretch}{0.7} % adjust vertical space
	%	\vspace{1.5cm}
	\begin{center}
		\begin{tabular}{ccccc}
			
			\includegraphics[width= 0.18\textwidth]{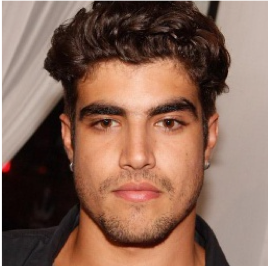}
			&
			\includegraphics[width= 0.18\textwidth]{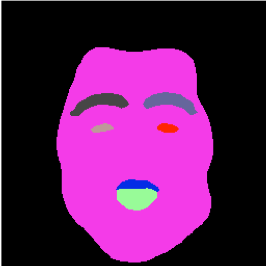}
			&
			\includegraphics[width= 0.18\textwidth]{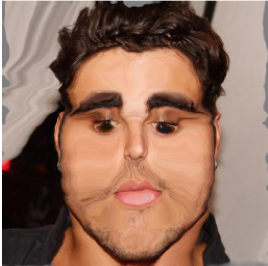}
			&
			\includegraphics[width=
			0.18\textwidth]{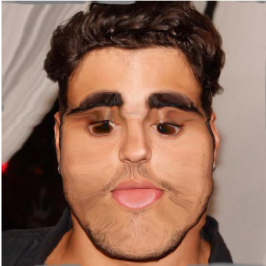}
			&
			\includegraphics[width=
			0.18\textwidth]{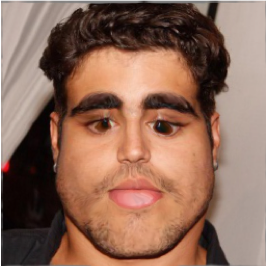}
			\\

			\includegraphics[width= 0.18\textwidth]{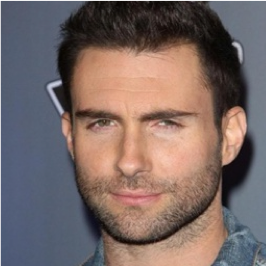}
			&
			\includegraphics[width= 0.18\textwidth]{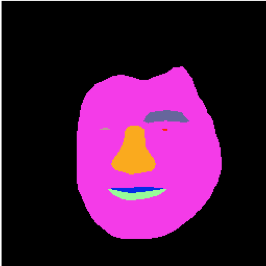}
			&
			\includegraphics[width= 0.18\textwidth]{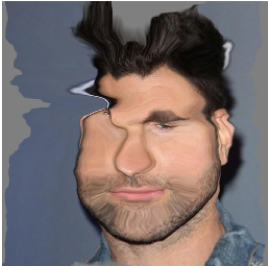}
			&
			\includegraphics[width=
			0.18\textwidth]{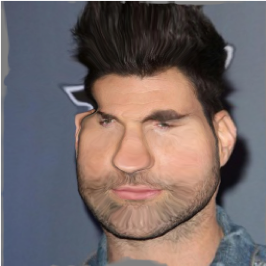}
			&
			\includegraphics[width=
			0.18\textwidth]{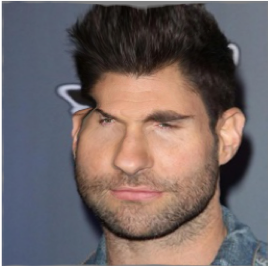}
			\\

			\addlinespace[0.1cm]
			Raw image &
			Parsing map&
			$+\mathcal{L}_{rec}$ &
			$+\mathcal{L}_{cyc}$ &
			$+\mathcal{L}_{coo}$
			
			\\
		\end{tabular}
		\vspace{-2mm}
	\end{center}
	
	\vspace{-0.1cm}
	\caption{{\bf Visual comparisons for loss functions}. 
	From left to right, we first show the raw image and then results of gradually adding $\mathcal{L}_{rec}$, $\mathcal{L}_{cyc}$, $\mathcal{L}_{coo}$.
	}
	\label{fig:ablation}
	\vspace{-2mm}
	
\end{figure*}

%HERE
%\vspace{-3mm}
%\paragraph{Coordinate-based Loss.}
{\flushleft \bf Coordinate-based Loss.}
The above-mentioned loss functions are all based on the parsing map $\mathbf{P}$ or regularization on $\mathbf{D}$, which constrain the output to be consistent in the semantic space.
Nevertheless, the constructed pixels may not be well aligned in the coordinate space.
To address this issue, we further introduce a coordinate-based loss when computing the cycle consistency.
Instead of only considering the parsing map $\mathbf{P}_{pho}$, we construct a coordinate map $\mathbf{M}_{pho} \in \mathbb{R}^{2 \times H \times W}$, where $\mathbf{M}_{pho}^{(i,j)} = (i, j) $ indicates the spatial location.
After obtaining the reconstructed $\mathbf{P}_{cyc}$, we convert it to a coordinate map $\mathbf{M}_{cyc}$.
Since this $\mathbf{M}_{cyc}$ has been operated through two decoders with the estimated warping parameters, the newly warped coordinates may not be aligned with $\mathbf{M}_{pho}$.
%
%MH: check this sentence
Thus we minimize the following loss in the coordinate space: 
\begin{equation}
    \mathcal{L}_{coo} = \frac{1}{H \times W} \sum_{i=1}^{H} \sum_{j=1}^{W} \|\mathbf{M}_{pho}^{(i,j)} - \mathbf{M}_{cyc}^{(i,j)}\|_{1}.
\end{equation}
%
%\red{
This spatially-variant consistency loss in the coordinate space can constrain per-pixel correspondence to be one-to-one and reversible, which reduces the artifacts inside each facial part.
%}
\subsection{Overall Objective}
% \yh{maybe we can reduce some minor loss functions to save the space}

The overall objective function for the proposed semantic shape transformation network includes the reconstruction/adversarial loss to help recover the semantic parsing map and the cycle consistency/coordinate-based loss to ensure consistency:
\begin{equation}
\begin{aligned}
    \mathcal{L}_{shape} = & \lambda_{r} \mathcal{L}_{rec} + \mathcal{L}_{adv} + \mathcal{L}_{cyc} + \mathcal{L}_{coo}.
\end{aligned}
\end{equation}
In this work, we regard the reconstruction term as a critical one and use $\lambda_{r} = 500$ in the following experiments.

\subsection{Implementation Details}
We implement our method using PyTorch~\citep{paszke2017automatic} and train the model with a single Nvidia 1080Ti GPU.
For the encoder, we 
utilize four basic dense blocks \citep{huang2017densely} to extract features, followed by a flatten layer to obtain a global $128$-d code for representing the semantic parsing map. 
The decoder consists of four symmetric dense blocks with transposed convolution layers for upsampling. 
During training, we utilize the Adam~\citep{kingma2014adam} optimizer and use the batch size as $32$.
Similar to CycleGAN, we set the initial learning rate as $0.0001$ and fix it for the first 300 epochs, and linearly decay the learning rate for another 300 epochs. 
%
%\red{
The source code of the proposed method is available at \url{https://github.com/wenqingchu/Semantic-CariGANs}.
%}

\section{Results and Analysis}

We evaluate the proposed algorithm and relevant methods on the large WebCaricature dataset~\citep{HuoBMVC2018WebCaricature}.
This photo-caricature benchmark dataset
contains $5974$ photos and $6042$ caricatures collected from the web. 
We use the provided landmarks to crop/align faces and then resize them into $256 \times 256$ pixels.
In addition, we randomly select $500$ photos as the test set and use the rest as the training set. 
We perform qualitative and quantitative experiments to demonstrate the effectiveness of the proposed algorithm.
%
%The source code and trained models will be made availalbe to the public. 

\subsection{Ablation Study on Shape Transformation}
We first analyze the quality of the proposed semantic shape transformation algorithm in this section.

\begin{table}[!t]
	
	\caption{{Ablation studies on different loss functions.}}
	%\vspace{-3mm}
	\label{table:comparison_loss}
	\scriptsize
	\newcommand{\tabincell}[2]{\begin{tabular}{@{}#1@{}}#2\end{tabular}}
	\newcolumntype{P}[1]{>{\centering\arraybackslash}p{#1}}
	
	\centering
	\begin{tabular}{m{3.0cm}  P{1.5cm} P{1.5cm}  }
		\toprule
		Loss functions & mIoU & pixAcc  \\
		\midrule
		w/o $\mathcal{L}_{rec}$ & 23.25  & 85.76   \\
		w/o $\mathcal{L}_{adv}$  & 58.19  & 94.66  \\
		w/o $\mathcal{L}_{cyc}$  & 58.42  & 94.70  \\
		w/o $\mathcal{L}_{coo}$   & 60.05  & 95.49  \\
		$\mathcal{L}_{shape}$ (ours)  & $\mathbf{61.66}$   & $\mathbf{95.60}$ \\
		%\midrule
		\bottomrule
	\end{tabular} 
	\vspace{-2mm}
\end{table}

\begin{figure*}
	
	\footnotesize
	%	\tiny
	\centering
	\renewcommand{\tabcolsep}{0.5pt} % adjust horizontal space
	\renewcommand{\arraystretch}{0.2} % adjust vertical space
	%	\vspace{1.5cm}
	\begin{center}
		\begin{tabular}{ccccc}
			%\multirow{2}{*}[0.96em]{	
			%	\includegraphics[width= %0.2\textwidth]{/images/examples/results_for_different_methods/CycleGan_1.png}
			%}
			%&   
			
			%	\vspace{-2pt}  
			\includegraphics[width= 0.18\textwidth]{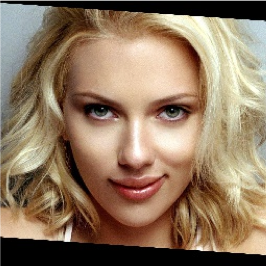}
			&
			\includegraphics[width= 0.18\textwidth]{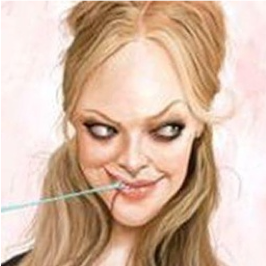}
			&
			\includegraphics[width= 0.18\textwidth]{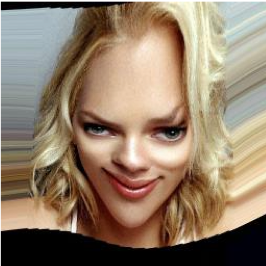}
			&
			\includegraphics[width= 0.18\textwidth]{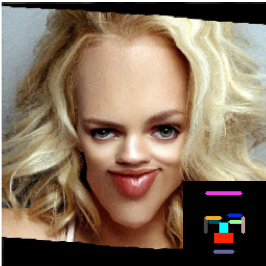}
			&
			\includegraphics[width= 0.18\textwidth]{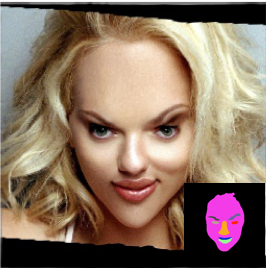}
			\\
			
			\includegraphics[width= 0.18\textwidth]{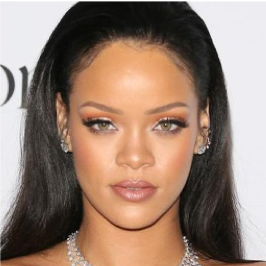}
			&
			\includegraphics[width= 0.18\textwidth]{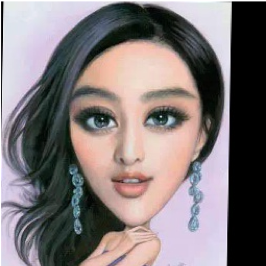}
			&
			\includegraphics[width= 0.18\textwidth]{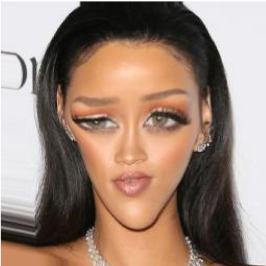}
			&
			\includegraphics[width= 0.18\textwidth]{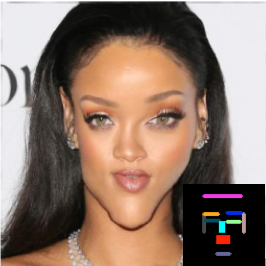}
			&
			\includegraphics[width= 0.18\textwidth]{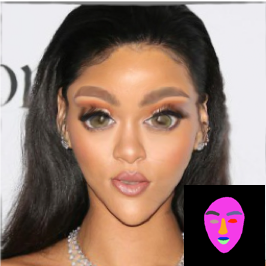}
			\\

			\includegraphics[width= 0.18\textwidth]{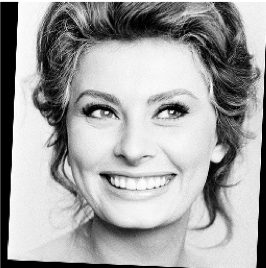}
			&
			\includegraphics[width= 0.18\textwidth]{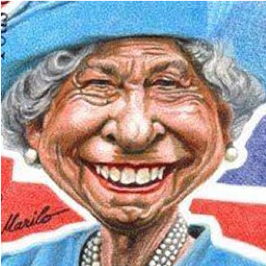}
			&
			\includegraphics[width= 0.18\textwidth]{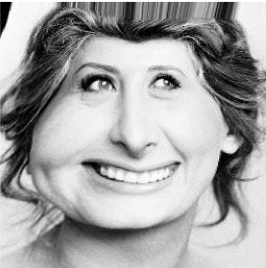}
			&
			\includegraphics[width= 0.18\textwidth]{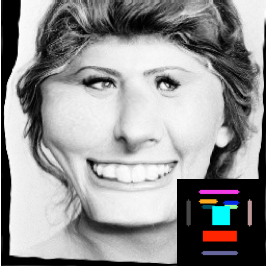}
			&
			\includegraphics[width= 0.18\textwidth]{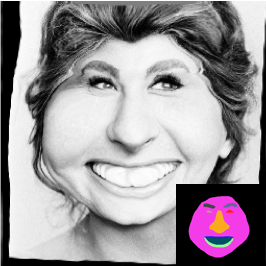}
			\\

 			\addlinespace[0.2cm]
			\scriptsize{Photo} &
			\scriptsize{Shape Reference} &
			\scriptsize{Landmark Positions} &
			\scriptsize{Landmark Maps} &
			\scriptsize{Parsing Maps (Ours)}
			\\
		\end{tabular}
		\vspace{-2mm}
	\end{center}

	\vspace{-0.2cm}
	\caption{{\bf Visual comparisons of shape transformation generated by different inputs.} Our method using face parsing maps is able to accurately transfer the shape exaggerations from the shape reference, while preserving the facial structure.}
	\label{fig:exp1}
	%\vspace{-8pt}
	
\end{figure*}

%\vspace{-3mm}
%\paragraph{Loss Functions.}
{\flushleft \bf Loss Functions.}
We evaluate the effectiveness of the proposed loss functions quantitatively. 
We randomly select 200 caricatures as reference images, guiding the test photos to generate transformed caricature-like output (without the style transfer stage).
Next, we utilize the parsing map of the reference caricature and verify whether the parsing map of the transformed photo is similar to the original parsing map in caricature.
We evaluate the performance with mean intersection-over-intersection (mIoU) and pixel accuracy (pixAcc), which are the common metrics for semantic segmentation. 

Table~\ref{table:comparison_loss} shows the results of this ablation study.
Without the reconstruction loss $\mathcal{L}_{rec}$, the model has no control to transform face shapes to be similar to the reference parsing map, resulting in the lowest accuracy.
Without the adversarial loss $\mathcal{L}_{adv}$ or cycle consistency loss $\mathcal{L}_{cyc}$, we observe that the training is less stable and is unable to preserve details.
Finally, although removing $\mathcal{L}_{coo}$ only slightly degrades the parsing mIoU, we notice that it is a crucial component to perform more accurate alignment for facial components.
%

% \wq{
In addition, we also provide visual comparisons in Fig.~\ref{fig:ablation} to verify the effectiveness of the designed loss functions qualitatively.
Here, we gradually add $\mathcal{L}_{rec}$, $\mathcal{L}_{cyc}$, and $\mathcal{L}_{coo}$ to our model.
We find that only using $\mathcal{L}_{rec}$ leads to severe artifacts, e.g., around the eye and skin regions. This is because this model only learns to change the semantic shapes and overlook the inherent structure of the facial component.
Introducing the additional $\mathcal{L}_{cyc}$ could constrain the mapping function and produce less distortion.
Finally, we observe that adding $\mathcal{L}_{coo}$ based on the coordinate space makes the constructed pixels well aligned and produces visually pleasing results. The reason is that $\mathcal{L}_{coo}$ penalizes pixels with large distortion and enforces all pixels in the same semantic region to have smoother translations.
% }

 %\vspace{-3mm}
%\paragraph{Comparisons to the Landmark Input.}
%\red{
{\flushleft \bf Comparisons to the Landmark Input.}
%}
%\wq{move this part from supp to 'results'}
In this work, we utilize the parsing maps as the input to learn shape transformation, while previous methods~\citep{cao2018carigans,li2018carigan} mainly leverage sparse facial landmarks for shape exaggeration.
We present an ablation study of using different inputs for performance evaluation. 
Similar to~\citep{cao2018carigans,li2018carigan}, we consider two approaches to make use of the labeled facial landmarks provided in the Webcaricature dataset and show visual comparisons in Fig.~\ref{fig:exp1}.

\begin{figure*}[!t]
	
	\footnotesize
	%	\tiny
	\centering
	\renewcommand{\tabcolsep}{0.5pt} % adjust horizontal space
	\renewcommand{\arraystretch}{0.2} % adjust vertical space
	%	\vspace{1.5cm}
	\begin{center}
		\begin{tabular}{cccccc}
			%\multirow{2}{*}[0.96em]{	
			%	\includegraphics[width= %0.2\textwidth]{/images/examples/results_for_different_methods/CycleGan_1.png}
			%}
			%&   
			
			%	\vspace{-2pt}  
			\includegraphics[width= 0.16\textwidth]{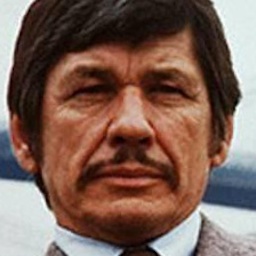}
			&
			\includegraphics[width= 0.16\textwidth]{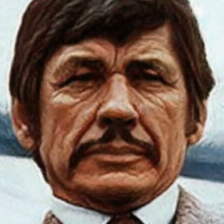}
			&
			\includegraphics[width= 0.16\textwidth]{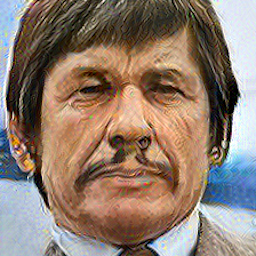}
			&
			\includegraphics[width= 0.16\textwidth]{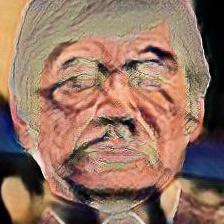}
			&
			\includegraphics[width= 0.16\textwidth]{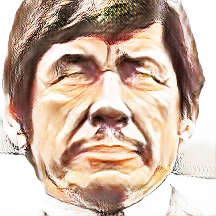}
			&
			\includegraphics[width= 0.16\textwidth]{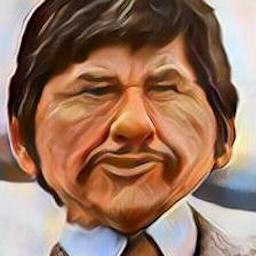}
			%\includegraphics[width=
			%0.12\textwidth]{images/comparisons/v1/pair08_ref.jpg}
			\\

			\includegraphics[width= 0.16\textwidth]{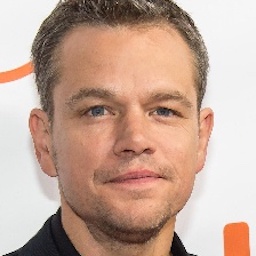}
			&
			\includegraphics[width= 0.16\textwidth]{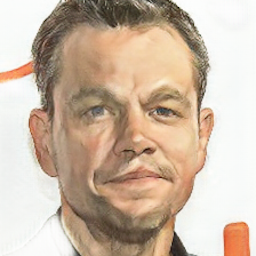}
			&
			\includegraphics[width= 0.16\textwidth]{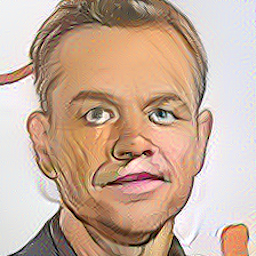}
			&
			\includegraphics[width= 0.16\textwidth]{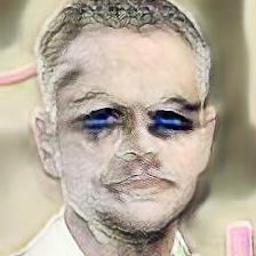}
			&
			\includegraphics[width= 0.16\textwidth]{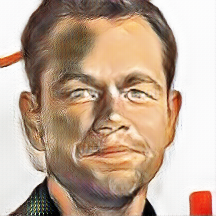}
			&
			\includegraphics[width= 0.16\textwidth]{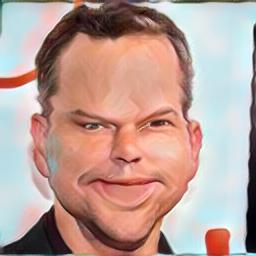}
			\\

  			\includegraphics[width= 0.16\textwidth]{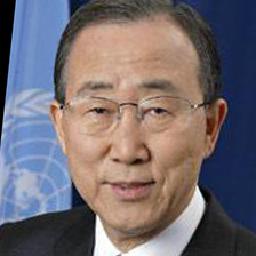}
  			&
  			\includegraphics[width= 0.16\textwidth]{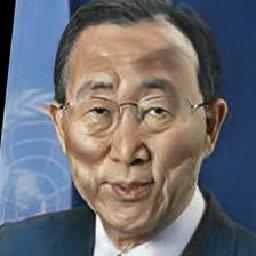}
  			&
  			\includegraphics[width= 0.16\textwidth]{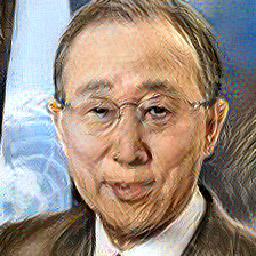}
  			&
  			\includegraphics[width= 0.16\textwidth]{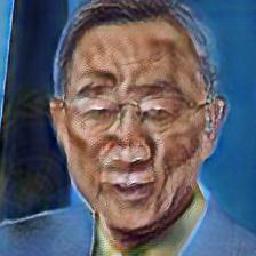}
  			&
  			\includegraphics[width= 0.16\textwidth]{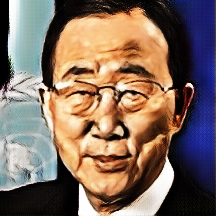}
  			&
  			\includegraphics[width= 0.16\textwidth]{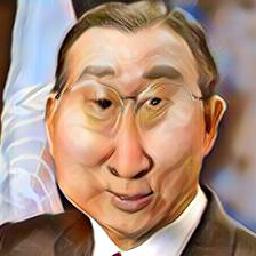}
  			\\

   			\includegraphics[width= 0.16\textwidth]{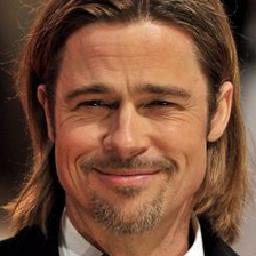}
   			&
   			\includegraphics[width= 0.16\textwidth]{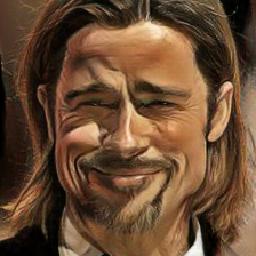}
   			&
   			\includegraphics[width= 0.16\textwidth]{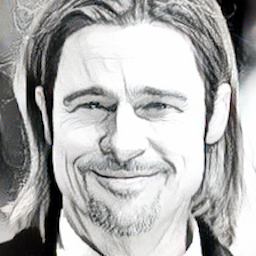}
   			&
   			\includegraphics[width= 0.16\textwidth]{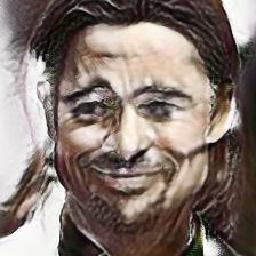}
   			&
   			\includegraphics[width= 0.16\textwidth]{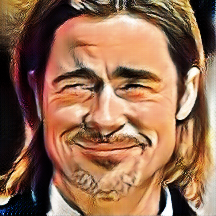}
   			&
   			\includegraphics[width= 0.16\textwidth]{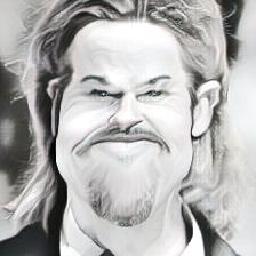}
   			\\			

   			\includegraphics[width= 0.16\textwidth]{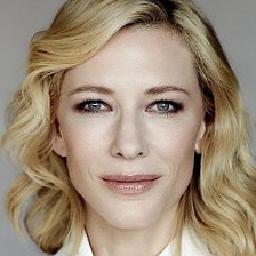}
   			&
   			\includegraphics[width= 0.16\textwidth]{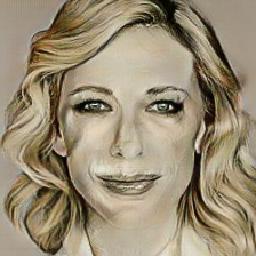}
   			&
   			\includegraphics[width= 0.16\textwidth]{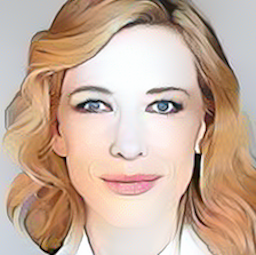}
   			&
   			\includegraphics[width= 0.16\textwidth]{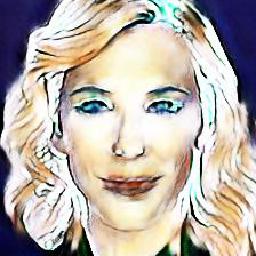}
   			&
   			\includegraphics[width= 0.16\textwidth]{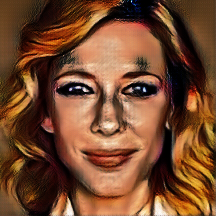}
   			&
   			\includegraphics[width= 0.16\textwidth]{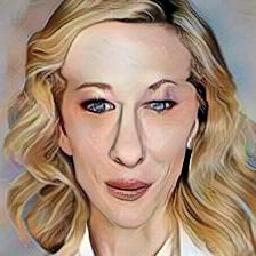}
   			\\	

			\addlinespace[0.1cm]
			\scriptsize{Photo} &
			\scriptsize{CycleGAN} &
			\scriptsize{Style Transfer} &
			\scriptsize{MUNIT} &
			\scriptsize{DRIT} &
			\scriptsize{Ours} \\
		\end{tabular}
		\vspace{-2mm}
	\end{center}

	\vspace{-0.2cm}
	\caption{
	%\yh{remove one row}
	Visual comparisons with different image translation methods.
% 	In each of our result, we show the shape reference on the bottom-right corner, while randomly choosing one style reference to generate the final output.
	}
	\label{fig:v1}
	\vspace{-2mm}
	
\end{figure*}

\begin{figure*}[!t]
	
	\footnotesize
	%	\tiny
	\centering
	\renewcommand{\tabcolsep}{0.5pt} % adjust horizontal space
	\renewcommand{\arraystretch}{0.2} % adjust vertical space
	%	\vspace{1.5cm}
	\begin{center}
		\begin{tabular}{cccccc}
			%\multirow{2}{*}[0.96em]{	
			%	\includegraphics[width= %0.2\textwidth]{/images/examples/results_for_different_methods/CycleGan_1.png}
			%}
			%&   

			\includegraphics[width= 0.16\textwidth]{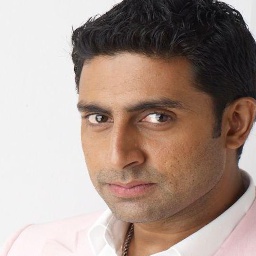}
			&
			\includegraphics[width= 0.16\textwidth]{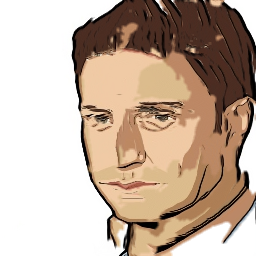}
			&
			\includegraphics[width= 0.16\textwidth]{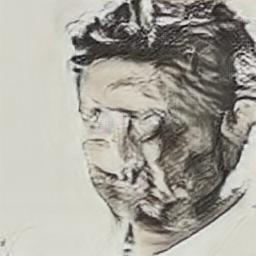}
			&
			\includegraphics[width= 0.16\textwidth]{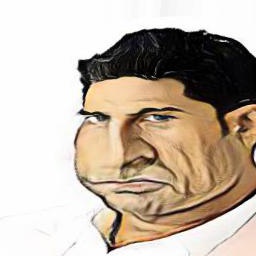}
			&
			\includegraphics[width=
			0.16\textwidth]{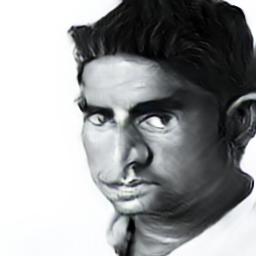}
			&
% 			\includegraphics[width= 0.16\textwidth]{images/comparisons/v2/100302_ours1.png}
% 			&
			\includegraphics[width= 0.16\textwidth]{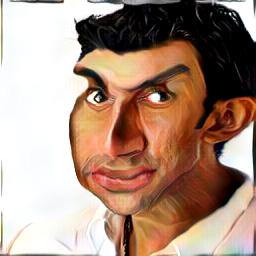}
			\\

			%\includegraphics[width= 0.12\textwidth]{images/comparisons/v2/pair16_input.jpg}
			%&
			%&
			%\includegraphics[width= 0.16\textwidth]{images/comparisons/v2/pair16_arxiv.jpg}
			%&
			%\includegraphics[width= 0.12\textwidth]{images/comparisons/v2/pair16_carigan.jpg}
			&
			%\includegraphics[width=
			%0.12\textwidth]{images/comparisons/v2/pair16_warpgan.jpg}
			%&
% 			\includegraphics[width= 0.16\textwidth]{images/comparisons/v2/092352_ours1.png}
% 			&
			%\includegraphics[width= 0.12\textwidth]{images/comparisons/v2/pair16_ours.jpg}
			%\includegraphics[width= 0.12\textwidth]{images/comparisons/v2/pair16_ref.jpg}
			\\				
			
			%	\vspace{-2pt}  
			
			\includegraphics[width= 0.16\textwidth]{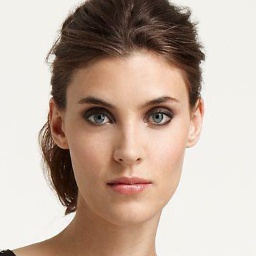}
			&
			\includegraphics[width= 0.16\textwidth]{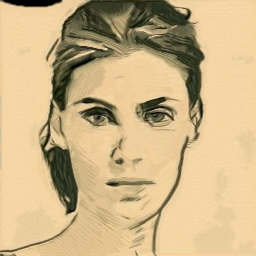}
			&
			\includegraphics[width= 0.16\textwidth]{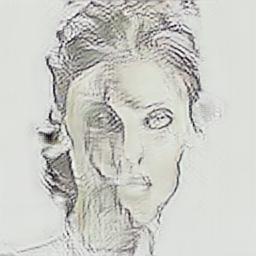}
			&
			\includegraphics[width= 0.16\textwidth]{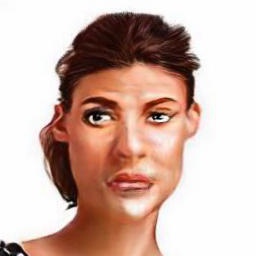}
			&
			\includegraphics[width=
			0.16\textwidth]{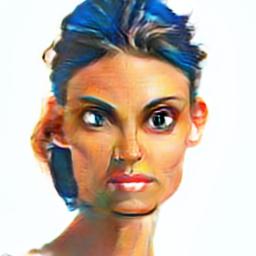}
			&
% 			\includegraphics[width= 0.16\textwidth]{images/comparisons/v2/067507_ours1.png}
% 			&
			\includegraphics[width= 0.16\textwidth]{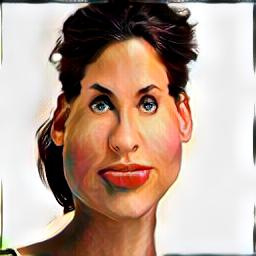}
			\\

 			\includegraphics[width= 0.16\textwidth]{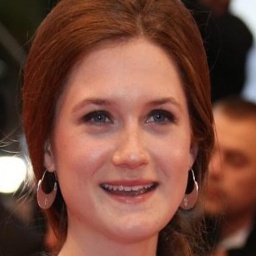}
 			&
 			\includegraphics[width= 0.16\textwidth]{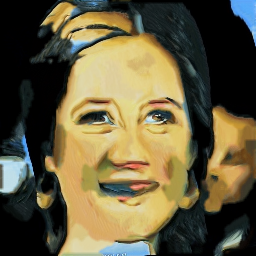}
 			&
 			\includegraphics[width= 0.16\textwidth]{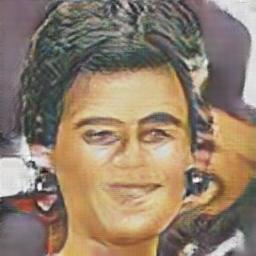}
 			&
 			\includegraphics[width= 0.16\textwidth]{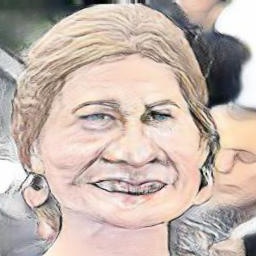}
 			&
 			\includegraphics[width= 0.16\textwidth]{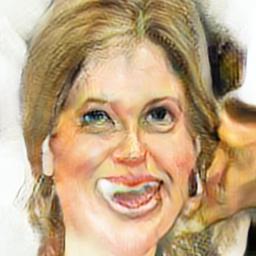}
 			&
 			\includegraphics[width= 0.16\textwidth]{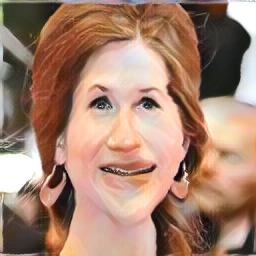}
 			\\

 			\includegraphics[width= 0.16\textwidth]{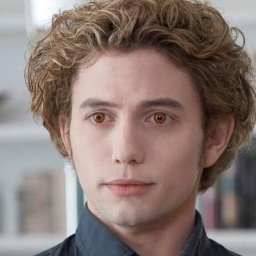}
 			&
 			\includegraphics[width= 0.16\textwidth]{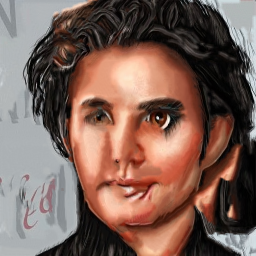}
 			&
 			\includegraphics[width= 0.16\textwidth]{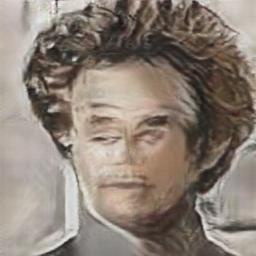}
 			&
 			\includegraphics[width= 0.16\textwidth]{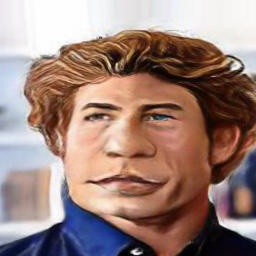}
 			&
 			\includegraphics[width= 0.16\textwidth]{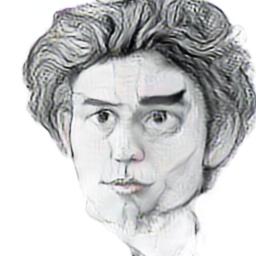}
 			&
 			\includegraphics[width= 0.16\textwidth]{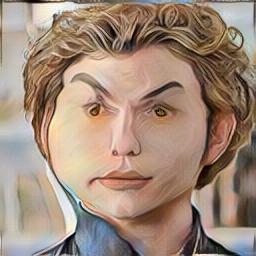}
 			\\

  			\includegraphics[width= 0.16\textwidth]{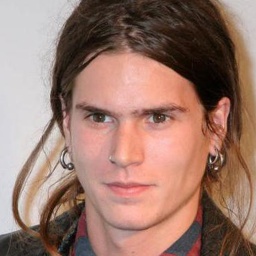}
  			&
  			\includegraphics[width= 0.16\textwidth]{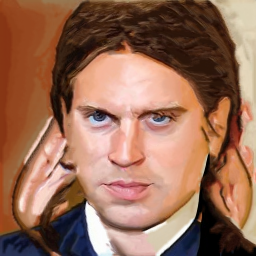}
  			&
  			\includegraphics[width= 0.16\textwidth]{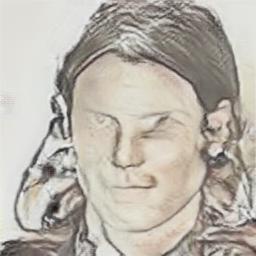}
  			&
  			\includegraphics[width= 0.16\textwidth]{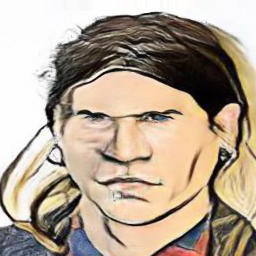}
  			&
 			\includegraphics[width= 0.16\textwidth]{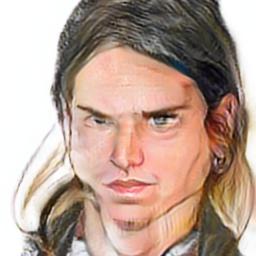}
 			&
  			\includegraphics[width= 0.16\textwidth]{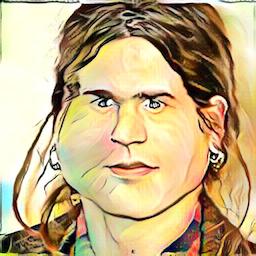}
  			\\

			\addlinespace[0.1cm]
			\scriptsize{Photo} &
			\scriptsize{Deep Analogy} &
		    \scriptsize{Zhang et al.} &
			\scriptsize{CariGANs} &
			\scriptsize{WarpGAN} &
% 			\scriptsize{Deep Analogy~\citep{liao2017visual}} &
% 		    \scriptsize{\cite{zheng2017photo}} &
% 			\scriptsize{CariGANs~\citep{cao2018carigans}} &
% 			\scriptsize{WarpGAN~\citep{shi2019warpgan}} &
			\scriptsize{Ours}
			
			\\
		\end{tabular}
		\vspace{-2mm}
	\end{center}

	\vspace{-0.2cm}
	\caption{
	%\yh{remove one row}
	Visual comparisons with different caricature generation methods.
% 	In each of our result, we show the shape reference on the bottom-right corner, while randomly choosing one style reference to generate the final output.
	}
	\label{fig:v2}
	\vspace{-2mm}
	
\end{figure*}

For the first method, we compute the warping parameters between the photo and the caricature using the thin-plate-spline transform~\citep{jaderberg2015spatial} by aligning the landmark positions.
We denote this baseline method as ``Landmark positions''.
Although landmarks are aligned, there are obvious distortions due to the lack of control points.
To further increase the control points, we connect 17 sparse landmarks that belong to the same semantic (e.g., both eyes) into a single polygon.
We then apply an one-hot encoding on the polygons that contain different facial components, in which this input is used for caricature generation as in the proposed model.
As shown in Fig.~\ref{fig:exp1}, although the results are visually more pleasing than the one using landmark positions, the facial contour does not transfer from caricatures to the outputs, since we have no control over the regions that are not covered by the landmark polygons.
In addition, we conduct a user study: among 15 participants, 89$\%$ of the total 310 votes prefer our results than the ones using landmarks.
Compared to the two baseline methods, the proposed algorithm uses face parsing maps to further control the shape transform with semantics and thus produce better results.

%\vspace{-3mm}
%\paragraph{Shape-based Methods.}
{\flushleft \bf Shape-based Methods.}
We present comparisons with a non-rigid registration method~\citep{jian2010robust} and the recently developed Neural Best-Buddies~\citep{aberman2018neural}.
Similar to Table~\ref{table:comparison_loss}, we evaluate the quality of the transformed parsing map. The mIoU and pixACC are (46.5\%, 84.7\%), (45.6\%, 61.7\%) for~\cite{jian2010robust} and~\cite{aberman2018neural} respectively, which are worse than ours as (61.7\%, 95.6\%).
One possible reason is that these methods cannot leverage semantic labels to guide registration, while ours is a learning-based model that considers semantics.

\subsection{Comparisons to the State of the Arts}
We conduct experiments with comparisons to the state-of-the-art methods by showing visual comparisons and performing user studies.
%

%\vspace{-3mm}
%\paragraph{Image Translation Methods.}
{\flushleft \bf Image Translation Methods.}
We evaluate our method with the state-of-the-art image translation algorithms, including CycleGAN~\citep{zhu2017unpaired}, Neural Style Transfer~\citep{gatys2016image}, MUNIT~\citep{huang2018multimodal}, and DRIT~\citep{DRIT}. 
For Neural Style Transfer, we randomly select 5 caricatures as style images.

As shown in Fig.~\ref{fig:v1}, conventional image translation methods are able to generate different textures in the generated outputs. 
However, there are severe artifacts due to the large texture variations in caricatures.
More importantly, these methods only render slight shape changes, which is unsatisfactory for caricature generation. 
\begin{figure*}
	
	\footnotesize
	%	\tiny
	\centering
	\renewcommand{\tabcolsep}{0.5pt} % adjust horizontal space
	\renewcommand{\arraystretch}{0.2} % adjust vertical space
	%	\vspace{1.5cm}
	\begin{center}
		\begin{tabular}{ccccccc}
            \includegraphics[width= 0.14\textwidth]{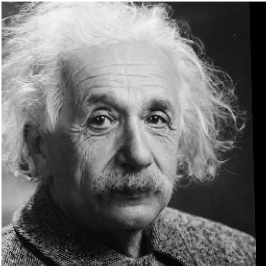}
			&
            \includegraphics[width= 0.14\textwidth]{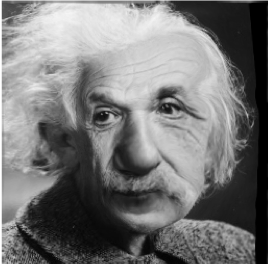}
			&
            \includegraphics[width= 0.14\textwidth]{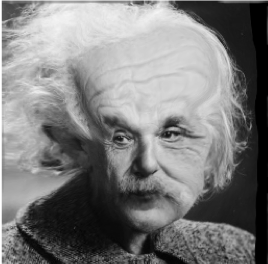}
			&
            \includegraphics[width= 0.14\textwidth]{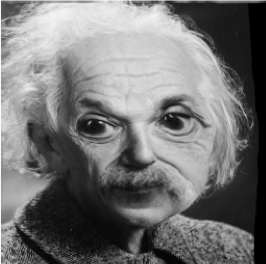}
			&
            \includegraphics[width= 0.14\textwidth]{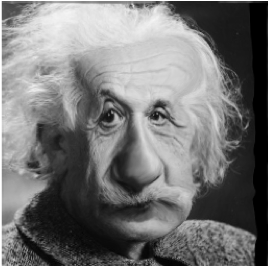}
			&
            \includegraphics[width= 0.14\textwidth]{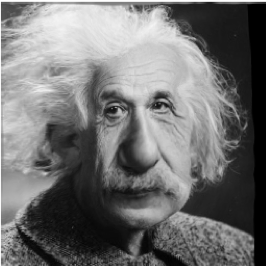}
			&
            \includegraphics[width= 0.14\textwidth]{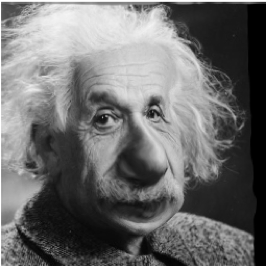}
		
			\\

            \includegraphics[width= 0.14\textwidth]{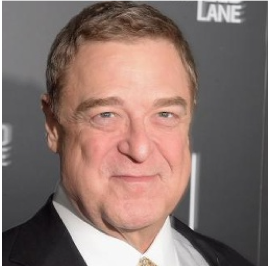}
			&
            \includegraphics[width= 0.14\textwidth]{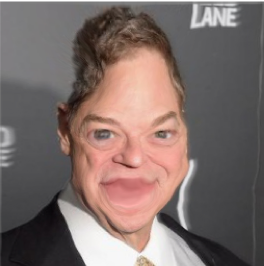}
			&
            \includegraphics[width= 0.14\textwidth]{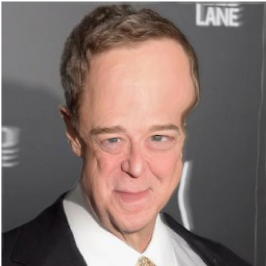}
			&
            \includegraphics[width= 0.14\textwidth]{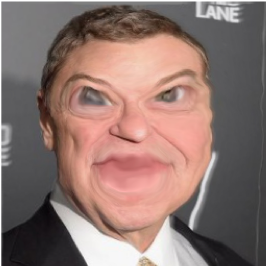}
			&
            \includegraphics[width= 0.14\textwidth]{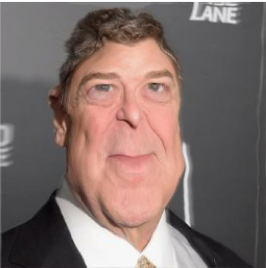}
			&
            \includegraphics[width= 0.14\textwidth]{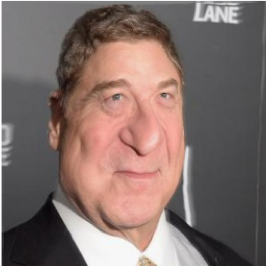}
			&
            \includegraphics[width= 0.14\textwidth]{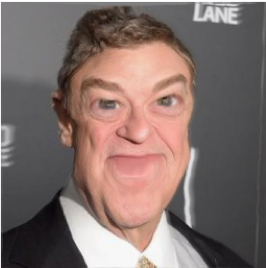}
		
			\\

            \includegraphics[width= 0.14\textwidth]{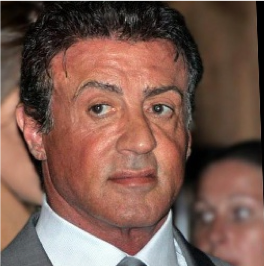}
			&
            \includegraphics[width= 0.14\textwidth]{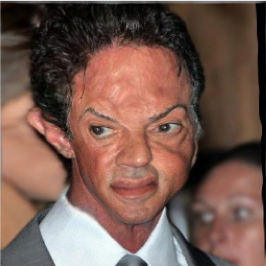}
			&
            \includegraphics[width= 0.14\textwidth]{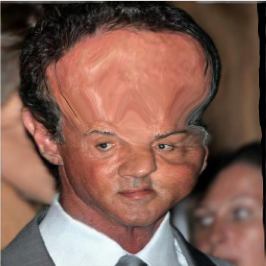}
			&
            \includegraphics[width= 0.14\textwidth]{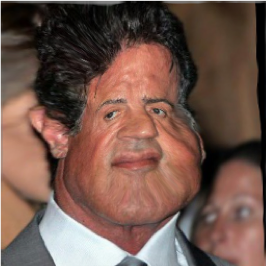}
			&
            \includegraphics[width= 0.14\textwidth]{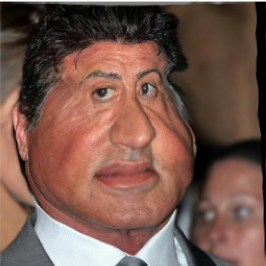}
			&
            \includegraphics[width= 0.14\textwidth]{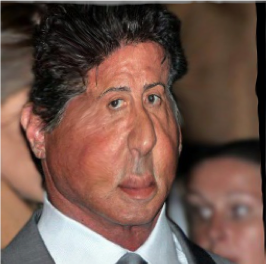}
			&
            \includegraphics[width= 0.14\textwidth]{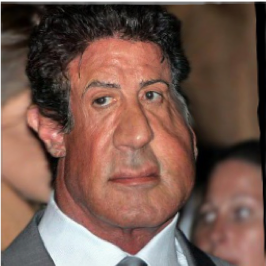}
		
			\\
		
 			\addlinespace[0.1cm]
			Photo &
			Random-1 &
			Random-2 &
			Random-3 &
			Retrieval-1 &
			Retrieval-2 &
			Retrieval-3
			\\
			\end{tabular}
			\vspace{-2mm}

	\end{center}

	\vspace{-0.2cm}
	\caption{{\bf Visual comparisons of deformed photos guided by random selected caricatures and the recommended ones.} Given a photo, our caricature retrieval model is able to automatically select plausible reference caricatures and obtain diverse results.}
	\label{fig:retrieval}
	\vspace{-2mm}
	%\vspace{-8pt}
	
\end{figure*}

%\vspace{-3mm}
%\paragraph{Caricature Generation Methods.}
{\flushleft \bf Caricature Generation Methods.}
Next, we perform the evaluation on existing caricature generation methods, including WarpGAN~\citep{shi2019warpgan}, CariGAN~\citep{cao2018carigans}, \cite{zheng2017photo} and Deep Image Analogy~\citep{liao2017visual}. 
We use the images provided by CariGAN as the test set for fair comparisons, where their output results are provided by the authors.
In Fig.~\ref{fig:v2}, while the other approaches show improved results compared to image translation baselines, the exaggerated shape effects may not be natural.
%
% \wq {
Note that WarpGAN does not need additional inputs, but it is less effective in aligning larger shape deformation.

 \begin{table}[!t]
 	\caption{User study on image translation methods, hand-drawn caricatures, and our method.}
 	%\vspace{-3mm}
 	\label{table:user_v1}
 	%\scriptsize
 	\small
 	\renewcommand{\arraystretch}{1}
     \setlength{\tabcolsep}{4pt}
 	%%\setlength{\abovecaptionskip}{0.cm}
 % 	\setlength{\belowcaptionskip}{-0.8cm}
 % 	\newcommand{\tabincell}[2]{\begin{tabular}{@{}#1@{}}#2\end{tabular}}
 %   \newcommand{\setTableWid}{ m{9pt}<{\centering} }
 % 	\newcolumntype{P}[1]{>{\centering\arraybackslash}p{#1}}
	
 	\centering
 	\begin{tabular}{ccccccc}
 		\toprule
 		Method & \tiny{CycleGAN} & \tiny{Style} & \tiny{MUNIT} & \tiny{DRIT} & \tiny{Hand-drawn} & \tiny{Ours} \\
 		\midrule
 		Score $\uparrow$  & \scriptsize{0.54} & \scriptsize{1.02} & \scriptsize{-0.82} & \scriptsize{0.75} & \scriptsize{2.32} & \scriptsize{2.02} \\
 		%\midrule
 		\bottomrule
 	\end{tabular} 
 	\vspace{-2mm}
	
 \end{table}

 \begin{table}[!t]
 	\caption{{User study on comparing our method with different caricature generation approaches.}}
 	%\vspace{-3mm}
 	\label{table:user_v2}
 	%\scriptsize
 	\small
 	\renewcommand{\arraystretch}{1}
     \setlength{\tabcolsep}{6pt}

 	\centering
 	\begin{tabular}{cccccc}
 		\toprule
 		Method & \tiny{Analogy} & \tiny{Zheng \etal} & \tiny{CariGANs} & \tiny{WarpGAN} & \tiny{Ours} \\
 		\midrule
 		Score $\uparrow$   & \scriptsize{0.13} & \scriptsize{-1.71} & \scriptsize{1.19} & \scriptsize{0.78} & \scriptsize{2.06} \\
 		%\midrule
 		\bottomrule
 	\end{tabular} 
 	\vspace{-2mm}
 \end{table}

%\vspace{-3mm}
%\paragraph{User Study.}
{\flushleft \bf User Study.}
We conduct user studies to evaluate the quality of generated caricatures in the above-mentioned methods.
For each subject, we first show a normal face with a few hand-drawn caricatures as the instruction to guide the users.
During the study, we randomly choose two of the methods among all the methods, and present one generated caricature for each method.
We then ask each subject to select the one that ``looks more like a caricature'' in terms of the shape exaggeration, artistic style, and image quality.
To compare with image translation methods, we collect 2650 pairwise results from a total of 38 participants, while for caricature generation algorithms, we collect 2220 pairwise outcomes from 31 participants.
In Table~\ref{table:user_v1} and Table~\ref{table:user_v2}, we show the normalized scores computed by the Bradley-Terry model~\citep{BradleyTerry}.
The results show that the proposed algorithm performs favorably against state-of-the-art methods.
In addition, compared with hand-drawn caricatures provided from the WebCaricature dataset~\citep{HuoBMVC2018WebCaricature} in Table~\ref{table:user_v1}, our score is closest to hand-drawn ones.

\begin{figure*}[!t]
	\footnotesize
	\centering
	\renewcommand{\tabcolsep}{1pt} % adjust horizontal space
	\renewcommand{\arraystretch}{0.7} % adjust vertical space
	\begin{tabular}{ccccccc}
		\includegraphics[width=0.12\linewidth]{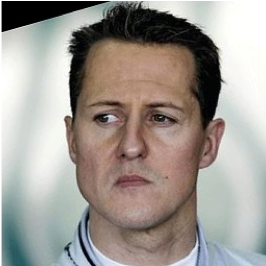} &
		\includegraphics[width=0.12\linewidth]{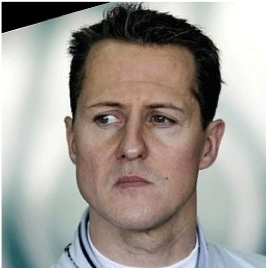} &
		\includegraphics[width=0.12\linewidth]{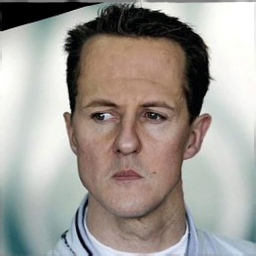} &
		\includegraphics[width=0.12\linewidth]{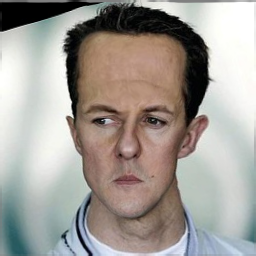} &
		\includegraphics[width=0.12\linewidth]{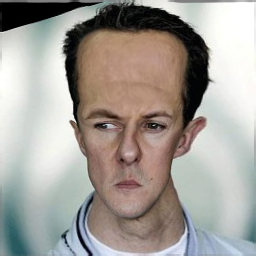} &
		\includegraphics[width=0.12\linewidth]{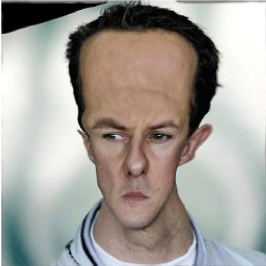} &
		\includegraphics[width=0.12\linewidth]{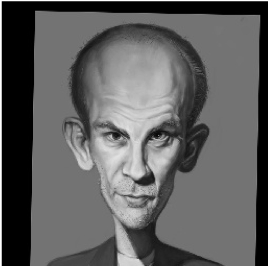} 
		\\

		\parbox[t]{2mm}{\multirow{1}{*}[3.3em]{\rotatebox[origin=c]{90}{}}} &
		\includegraphics[width=0.12\linewidth]{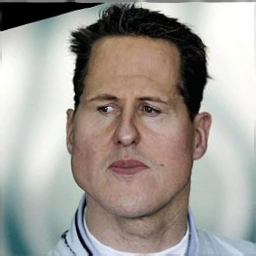} &
		\includegraphics[width=0.12\linewidth]{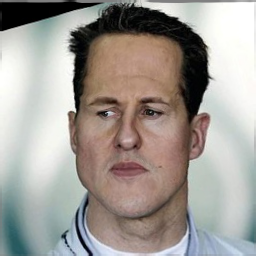} &
		\includegraphics[width=0.12\linewidth]{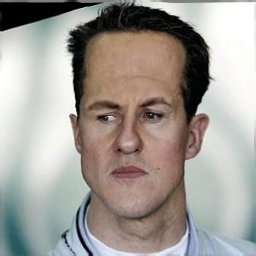} &
		\includegraphics[width=0.12\linewidth]{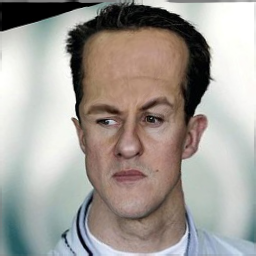} &
		\includegraphics[width=0.12\linewidth]{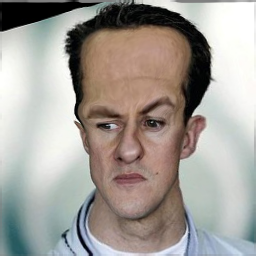} &
		\\

		\parbox[t]{2mm}{\multirow{1}{*}[3.3em]{\rotatebox[origin=c]{90}{}}} &
		\includegraphics[width=0.12\linewidth]{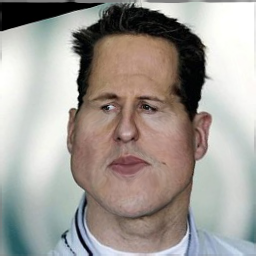} &
		\includegraphics[width=0.12\linewidth]{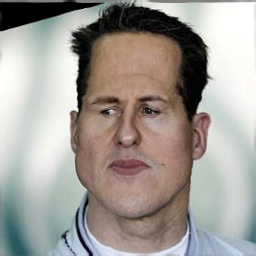} &
		\includegraphics[width=0.12\linewidth]{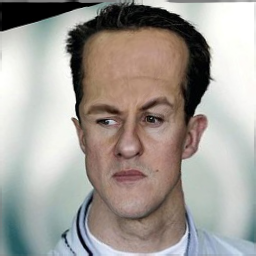} &
		\includegraphics[width=0.12\linewidth]{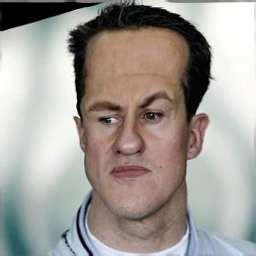} &
		\includegraphics[width=0.12\linewidth]{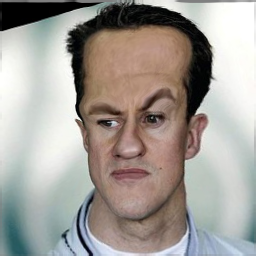} &
		\\

		\includegraphics[width=0.12\linewidth]{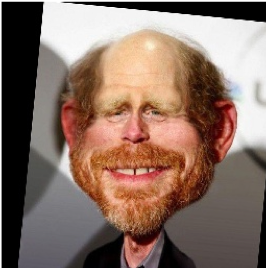} &
		\includegraphics[width=0.12\linewidth]{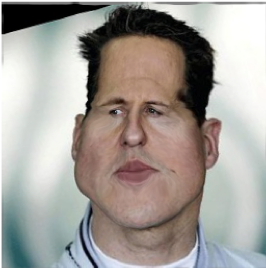} &
		\includegraphics[width=0.12\linewidth]{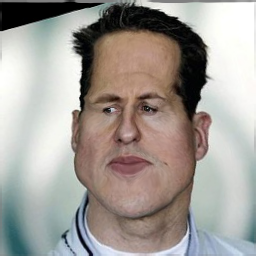} &
		\includegraphics[width=0.12\linewidth]{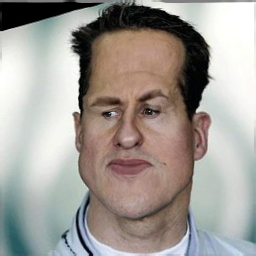} &
		\includegraphics[width=0.12\linewidth]{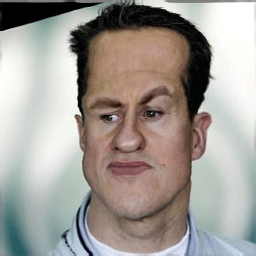} &
		\includegraphics[width=0.12\linewidth]{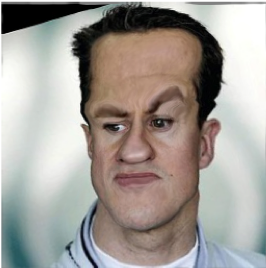} &
		\includegraphics[width=0.12\linewidth]{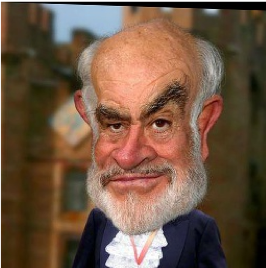} 
		\\
		
	\end{tabular}
	\vspace{-2mm}
	\caption{
% 	\yh{maybe remove one row and one column to make it look better}
	Shape interpolations between the input image and caricature in our learned shape encoding space.
% 	Due to this property, our model is allowed for users to interpolate the shape and select the degree of shape transformation via a simple visualization way.
	} 
	\label{fig:interpolation} %% label for entire figure
	\vspace{-2mm}
\end{figure*}

%\vspace{-3mm}
\subsection{Additional Results and Analysis}
%

%\paragraph{Caricature Retrieval.}
%\red{
%{\flushleft \bf Caricature Retrieval.}
%}
%\wq{move it from supp to 'results'}
Given a photo, our caricature retrieval model is able to automatically select plausible reference caricatures and obtain diverse results (Fig.~\ref{fig:teaser}). 
To demonstrate the effectiveness of this recommendation, we also select reference caricatures by random selection and compare the generated results through a user study. 
Specifically, we show two sets of deformed photos and allow users to select the preferred set based on the visual quality and diversity of generated caricatures.
Among 900 pairwise outcomes from 30 participants, 83$\%$ of the votes prefer our results. 
Here, we show more results of deformed photos guided by random selected caricatures and the recommended ones in Fig.~\ref{fig:retrieval}.
For simplicity, we only present results of deformed photos before applying style transfer.

\begin{figure}[!t]
	\begin{center}
		%\fbox{\rule{0pt}{3in} \rule{0.9\linewidth}{0pt}}
		\includegraphics[width=0.9\linewidth]{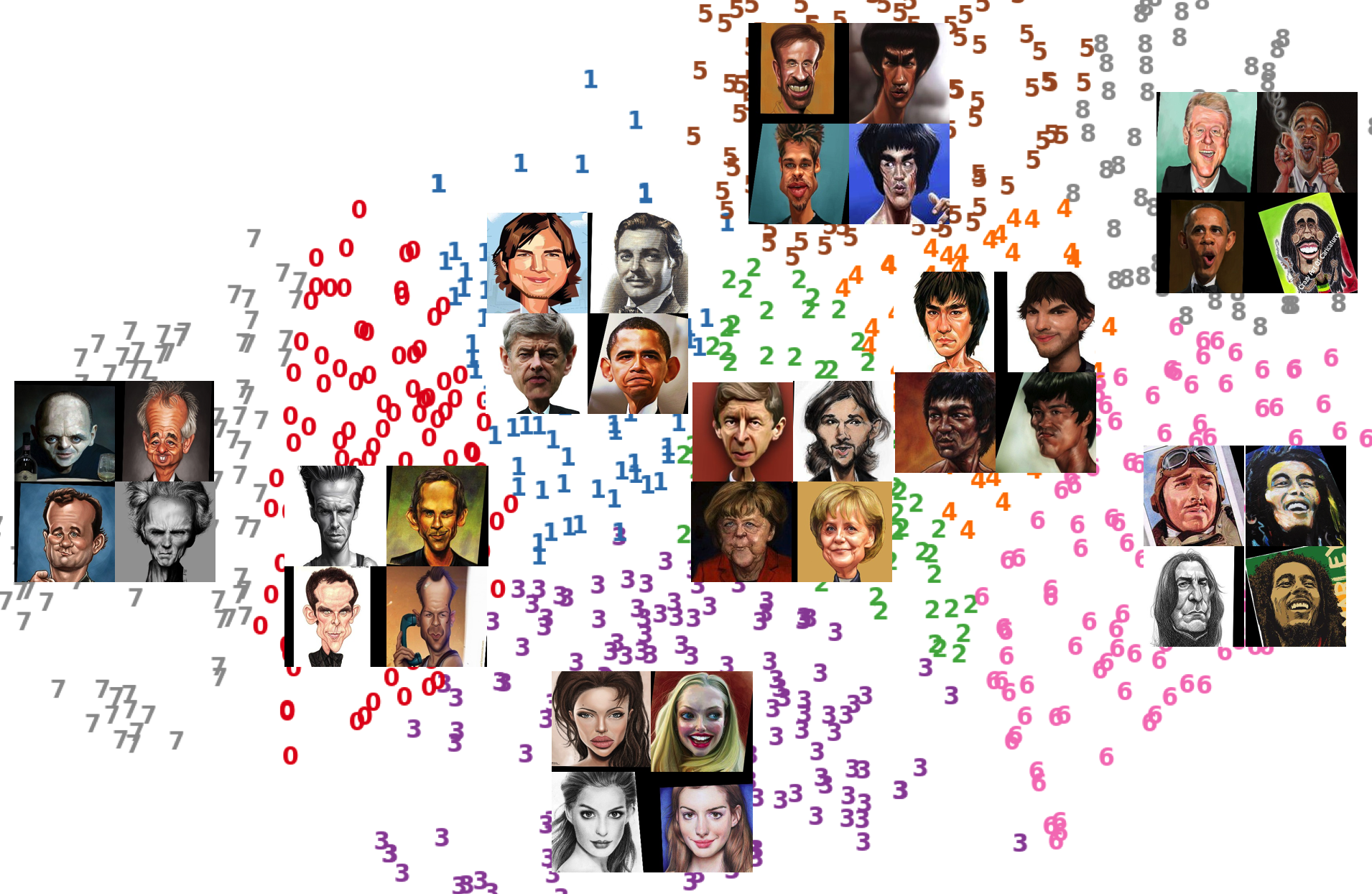}
	\end{center}
	\vspace{-2mm}
	\caption{
	{\bf Visualization of the encoding space for caricature shapes.} We show that each group (denoted in different colors) has certain shapes in facial components, while neighbors in this feature space also share similar shapes.}
	\label{fig:tsne}
	\vspace{-2mm}
\end{figure}

%\paragraph{Shape Embedding Space.}
{\flushleft \bf Shape Embedding Space.}
\label{sec:embedding}
We analyze the latent shape embedding space extracted from the encoder.  
We randomly select 1000 caricatures and extract their shape embedding vectors.
To verify whether the shape embedding features could capture meaningful facial structure information, we first apply the mean shift clustering method \citep{meanshift} to group caricature shapes and then apply the t-SNE \citep{maaten2008visualizing} scheme for visualization.
Fig.~\ref{fig:tsne} shows that caricatures within the same cluster share a similar facial structure, while neighboring clusters are also similar to each other in certain semantic parts.

We note that the embedding space shows smooth transitions between different shape deformation clusters, which can be exploited to generate different caricatures through a simple interpolation between caricature references as shown in Fig.~\ref{fig:interpolation}.
As a result, the degree of shape exaggeration could be controlled through interpolating between the input facial shape and the reference one, whenever users prefer to preserve more identity of the input portrait.

\begin{figure}[!t]
	\footnotesize
	\centering
 	\renewcommand{\tabcolsep}{1pt} % adjust horizontal space
 	\renewcommand{\arraystretch}{0.7} % adjust vertical space
	\begin{tabular}{cccc}

		\includegraphics[width=0.235\linewidth]{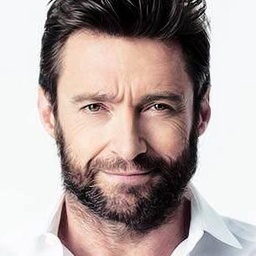} &

 		\includegraphics[width=0.235\linewidth]{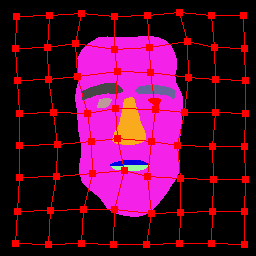}&
 		\includegraphics[width=0.235\linewidth]{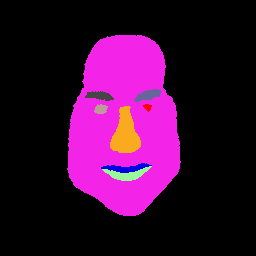}&
 		\includegraphics[width=0.235\linewidth]{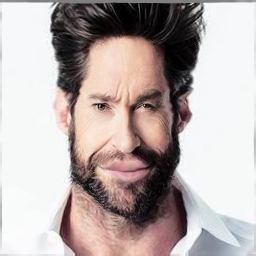}
 		\\

 		\parbox[t]{2mm}{\multirow{1}{*}[3.3em]{\rotatebox[origin=c]{90}{}}} &
		\includegraphics[width=0.235\linewidth]{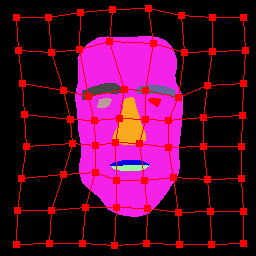}&
		\includegraphics[width=0.235\linewidth]{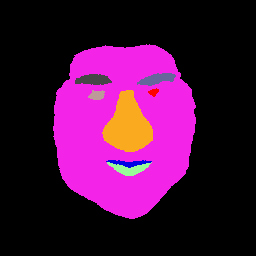}&
		\includegraphics[width=0.235\linewidth]{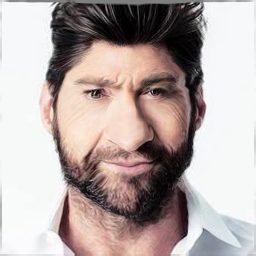}
		\\
		
		& Grid Controls & Edited Map & Deformed Photo \\
		
	\end{tabular}
	\vspace{-2mm}
	\caption{{\bf Results of manipulating on the parsing map.}
% 	We allow the users to control the grids and thereby obtain an updated parsing map to generate preferred shapes in caricature generation.
	Here, for simplicity, we only present results of deformed photos before applying style transfer.
	}
	\label{fig:control} %% label for entire figure
	\vspace{-2mm}
\end{figure}
%\paragraph{User Control.}
{\flushleft \bf User Control.}
Given a caricature/photo pair, our approach is not only as simple as a one-click shape transformation, but is also flexible to accommodate fine-grained facial structure refinements from users via providing gird controls on the parsing map.
We show an example for results controlled by the grid in Fig.~\ref{fig:control}.
Through adjusting the positions of control points, we are able to change the face contour, move the facial component positions, or adjust the shape of facial components. Therefore, users could manipulate the desired shape easily and create diverse caricatures. 
%
%\red{
In addition, these results demonstrate that it is plausible to feed new caricature parsing maps into our shape transformation model, e.g., the parsing map could be flexibly modified by the users as shown in Fig.~\ref{fig:control}.
%}
% In addition, as mentioned in Section~\ref{sec:embedding}, each user can adjust the trade-off parameter by interpolating between caricature and photo shape representations to achieve proper effect as shown in Fig.~\ref{fig:interpolation}. 

 %\vspace{-3mm}
 %\paragraph{Identity Preservation.}
 {\flushleft \bf Identity Preservation.}
In addition, we conduct a user study similar to CariGAN \citep{cao2018carigans} to evaluate the degree of identity preservation.
The users are asked to choose the correct subject from 5 portraits, given the results generated by each method.
We collect 650 votes with 26 participants and the accuracy are 70\% (our method), 53\% CariGAN \citep{cao2018carigans}, 
53\% \citep{zheng2017photo}), 68\% WarpGAN \citep{shi2019warpgan}, and 45\% Deep Image Analogy \citep{liao2017visual}.
%as shown in Table~\ref{table:user_v3}.
%
The results show that our model is the best one to preserve the identity.

\begin{figure*}[!t]
    	
    	\footnotesize
    	%	\tiny
    	\centering
    	\renewcommand{\tabcolsep}{0.5pt} % adjust horizontal space
    	\renewcommand{\arraystretch}{0.2} % adjust vertical space
    	%	\vspace{1.5cm}
    	\begin{center}
    		\begin{tabular}{ccccc}

    			\includegraphics[width= 0.16\textwidth]{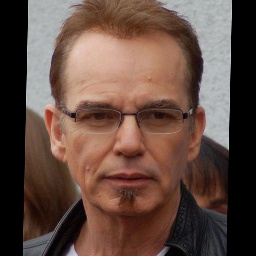}
    			&
    			\includegraphics[width= 0.16\textwidth]{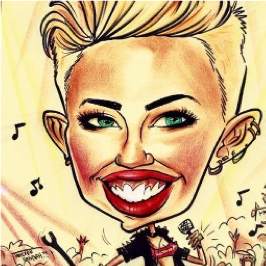}
    			&
    			\includegraphics[width= 0.16\textwidth]{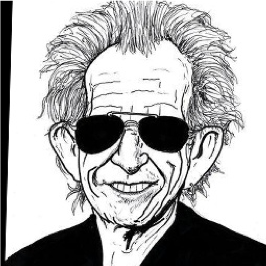}
    			&
    			\includegraphics[width= 0.16\textwidth]{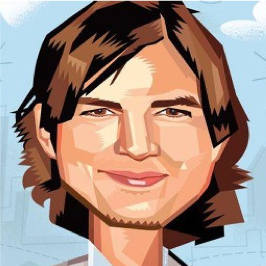}
    			&

    			\includegraphics[width= 0.16\textwidth]{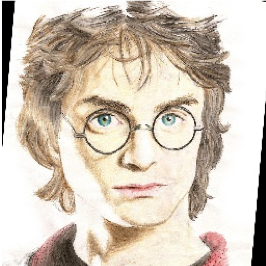}
    			\\

    			\includegraphics[width= 0.16\textwidth]{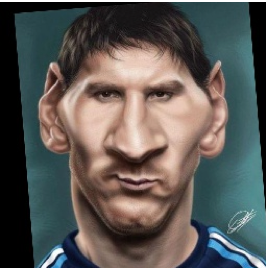}
    			&
    			\includegraphics[width= 0.16\textwidth]{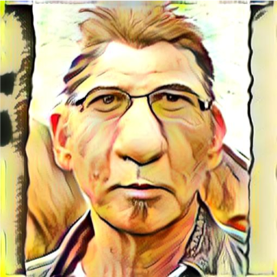}
    			&
    			\includegraphics[width= 0.16\textwidth]{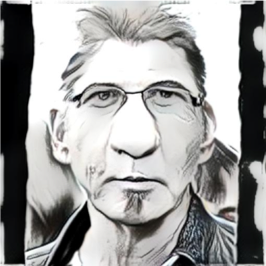}
    			&
    			\includegraphics[width= 0.16\textwidth]{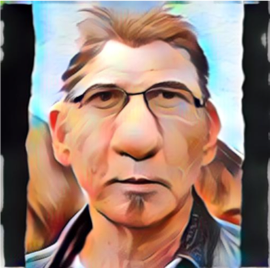}
    			&

    			\includegraphics[width= 0.16\textwidth]{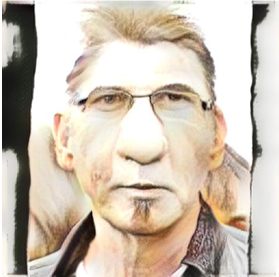}
    			\\

    			\includegraphics[width= 0.16\textwidth]{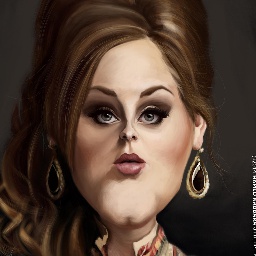}
    			&
    			\includegraphics[width= 0.16\textwidth]{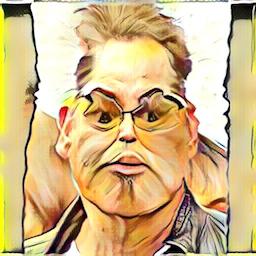}
    			&
    			\includegraphics[width= 0.16\textwidth]{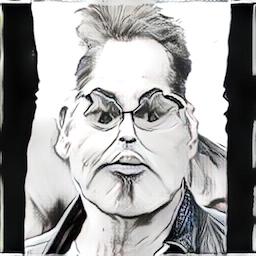}
    			&
    			\includegraphics[width= 0.16\textwidth]{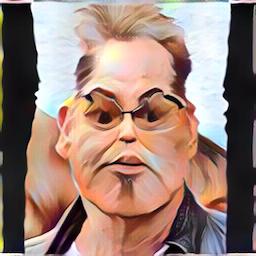}
    			&

    			\includegraphics[width= 0.16\textwidth]{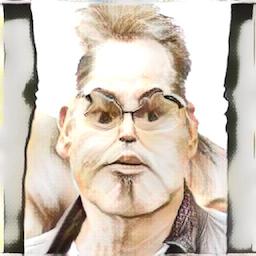}
    			\\

    			\includegraphics[width= 0.16\textwidth]{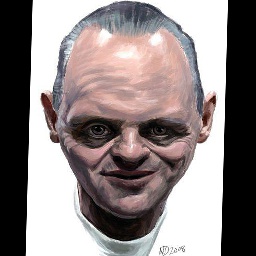}
    			&
    			\includegraphics[width= 0.16\textwidth]{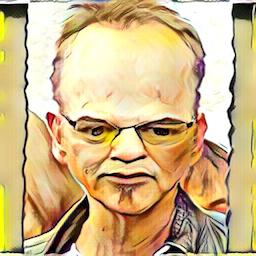}
    			&
    			\includegraphics[width= 0.16\textwidth]{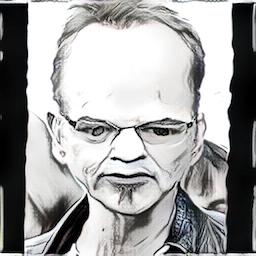}
    			&
    			\includegraphics[width= 0.16\textwidth]{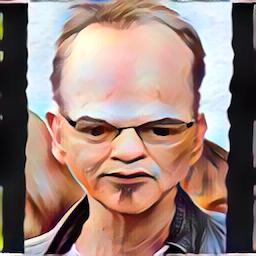}
    			&
    			\includegraphics[width= 0.16\textwidth]{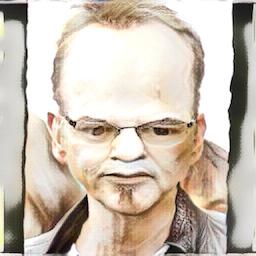}
    			\\ 	      
   
    			\includegraphics[width= 0.16\textwidth]{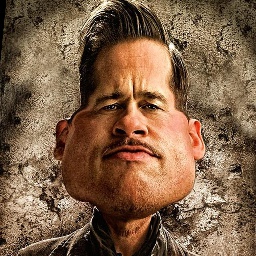}
    			&
    			\includegraphics[width= 0.16\textwidth]{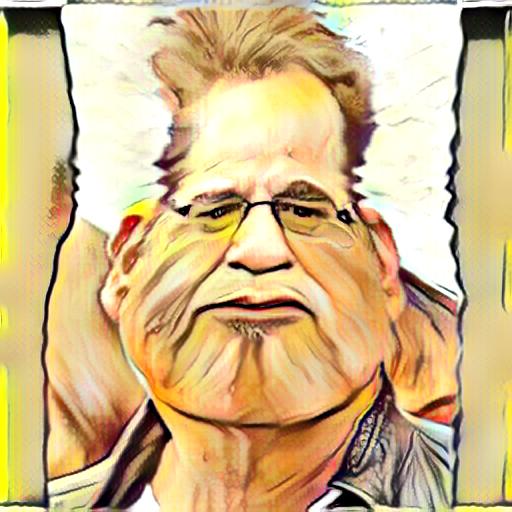}
    			&
    			\includegraphics[width= 0.16\textwidth]{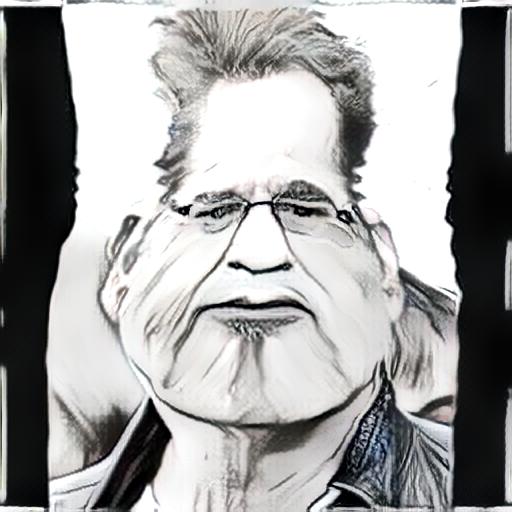}
    			&
    			\includegraphics[width= 0.16\textwidth]{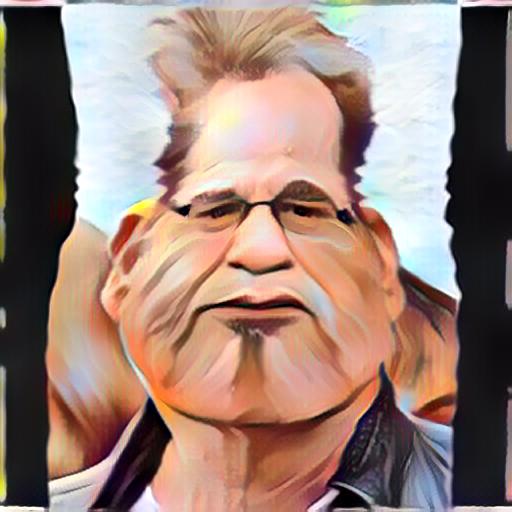}
    			&
    			\includegraphics[width= 0.16\textwidth]{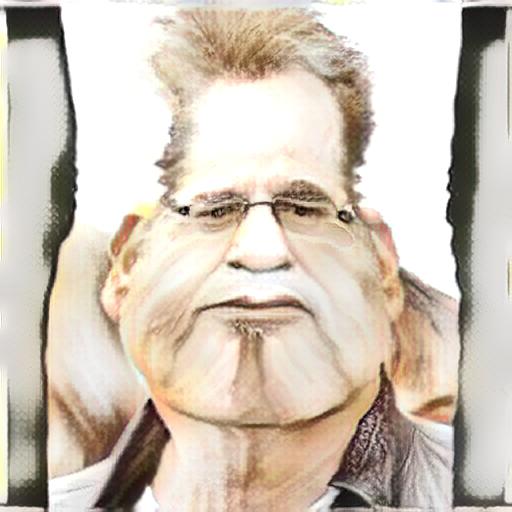}
    			
    			\\         
    		\end{tabular}
    	\end{center}

    	\vspace{-0.2cm}
    	\caption{{\bf Diverse results from different combinations for styles and caricatures.} The top-left image is the input photo. The first row is the caricatures for style reference, while the left-most column is the caricatures for shape reference.}
    	\label{fig:reference}
    	% 	\vspace{-8pt}
    	
    \end{figure*}

{\flushleft \bf Results with Diverse Styles.}
%\paragraph{Results with Diverse Styles.} 
The proposed algorithm utilize caricatures as reference images to guide the shape transformation and style transfer. In Fig.~\ref{fig:reference}, we show the input photo in the top-left corner. In the first row, we show reference caricatures for style transfer, while in the left-most column, we show the reference caricatures of the preferred shape.
With various combinations, our method is able to robustly produce diverse results with different styles and shapes.

\begin{figure}[!t]
 	\footnotesize
 	\centering
 	\renewcommand{\tabcolsep}{1pt} 
 	% adjust horizontal space
 	\renewcommand{\arraystretch}{0.7} 
 	% adjust vertical space
 	\begin{tabular}{cccc}
		
 		\includegraphics[width=0.235\linewidth]{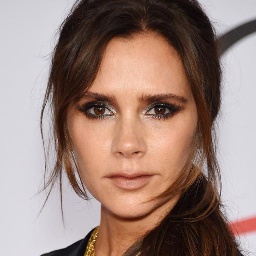} &
 		\includegraphics[width=0.235\linewidth]{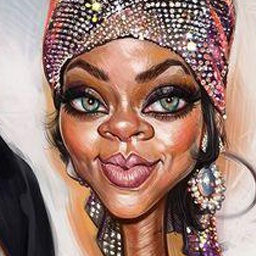} &
 		\includegraphics[width=0.235\linewidth]{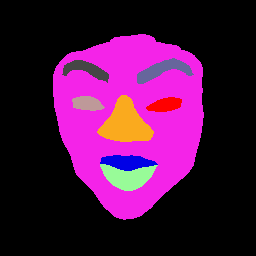} &
 		\includegraphics[width=0.235\linewidth]{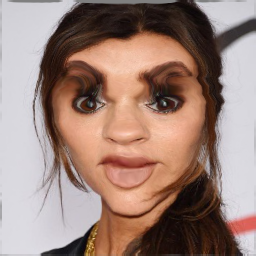}		
 		\\
		
 		Photo & Caricature & Parsing Map & Output \\
		
 	\end{tabular}
 	\vspace{-2mm}
 	\caption{Limitation of the proposed method.} 
 	\label{fig:limitation} %% label for entire figure
 	%\vspace{-12pt}
 	\vspace{-2mm}
 \end{figure}

%\vspace{-3mm}
%\paragraph{Limitation.}
{\flushleft \bf Limitation.}
Although using the proposed shape transformation for caricature generation is simple and effective, we find that it may render unsatisfactory results for some scenarios. 
For example, when the eyebrows and eyes are very close to each other in the photo, there would be some artifacts around the eyes when the transformer tries to change the eyebrow shapes as shown in Fig.~\ref{fig:limitation}. 
In the future work, we plan to use multiple shape sub-transformers for each facial component, followed by a global refinement network to integrate different sub-transformers together.

%\vspace{-3mm}
%\paragraph{Runtime Performance.}
{\flushleft \bf Runtime Performance.}
In the proposed framework, we use a single Nvidia 1080Ti GPU and the runtime on a $256 \times 256$ input photo is around 0.65 seconds, including 0.1 seconds for face parsing, 0.1 seconds for caricature retrieval, 0.3 seconds for shape transformation, and 0.15 seconds for style transfer.

\section{Conclusions}

In this paper, we propose a semantic dense shape transformation algorithm for learning to caricature.
%
%In order to perform the shape and style domain shifts, the input photo is first fed into a shape transformation network, followed by a style transfer process.
%
%For shape transformation, we utilize a face parsing map to densely predict warping parameters, such that the shape exaggerations are effectively transferred while the facial structure is still maintained.
Specifically, we utilize a face parsing map to densely predict warping parameters, such that the shape exaggerations are effectively transferred while the facial structure is still maintained.
Visual comparisons and user studies demonstrate that the proposed algorithm is able to generate high-quality caricatures against state-of-the-art methods.
In addition, we show that the learned embedding space of semantic parsing map allows us to directly manipulate the parsing map and generate shape changes according to the user preference.
% Although in this work we only evaluate our approach on caricature generation, it is not limited to the specific task of image translation.
% %
% The proposed method can be extended to nume
rous challenging image translation problems which require complex shape deformation.

% DO NOT INCLUDE ACKNOWLEDGMENTS IN AN ANONYMOUS SUBMISSION TO SIGGRAPH 2019
%\begin{acks}
%
%The authors would like to thank Dr. Maura Turolla of Telecom
%Italia for providing specifications about the application scenario.
%
%The work is supported by the \grantsponsor{GS501100001809}{National
%  Natural Science Foundation of
%  China}{http://dx.doi.org/10.13039/501100001809} under Grant
%No.:~\grantnum{GS501100001809}{61273304\_a}
%and~\grantnum[http://www.nnsf.cn/youngscientists]{GS501100001809}{Young
%  Scientists' Support Program}.
%
%
%\end{acks}

%   \section*{Acknowledgement}
%   This work is partially supported by National Science and Technology Major Project (No. 2018ZX01008103), NSF CARRER (No.1149783), and gifts from Adobe and Nvidia.
  
{\small
\bibliographystyle{spbasic}
\bibliography{caricature-bibliography}
}
  
  \end{document}